\documentclass[conference,compsoc]{IEEEtran}

\usepackage{epsfig}
\usepackage{float}
\usepackage{graphicx}

\ifCLASSOPTIONcompsoc
  \usepackage[nocompress]{cite}
\else
  \usepackage{cite}
\fi
\ifCLASSINFOpdf
\else
\fi
\usepackage{xcolor}
\usepackage{hyphenat}
\usepackage[breaklinks]{hyperref}
\usepackage{parskip}

\begin{document}
%
\title{All You Need In Sign Language Production}


\author{Razieh Rastgoo$^{1}$, Kourosh Kiani$^1$, Sergio Escalera$^2$, Vassilis Athitsos$^3$, Mohammad Sabokrou$^4$  \\
$^1$Semnan University~~ $^2$Universitat de Barcelona and Computer Vision Center\\
~~$^3$University of Texas at Arlington
~~$^4$Institute for Research in Fundamental Sciences (IPM)}
\maketitle

\begin{abstract}
Sign Language is the dominant form of communication language used in the deaf and hearing-impaired community. To make an easy and mutual communication between the hearing-impaired and the hearing communities, building a robust system capable of translating the spoken language into sign language and vice versa is fundamental. To this end, sign language recognition and production are two necessary parts for making such a two-way system. Sign language recognition and production need to cope with some critical challenges. In this survey, we review recent advances in Sign Language Production (SLP) and related areas using deep learning. To have more realistic perspectives to sign language, we present an introduction to the Deaf culture, Deaf centers, psychological perspective of sign language, the main differences between spoken language and sign language. Furthermore, we present the fundamental components of a bi-directional sign language translation system, discussing the main challenges in this area. Also, the backbone architectures and methods in SLP are briefly introduced and the proposed taxonomy on SLP is presented. Finally, a general framework for SLP and performance evaluation, and also a discussion on the recent developments, advantages, and limitations in SLP, commenting on possible lines for future research are presented.
\end{abstract}

\section{Introduction}
\begin{figure}[t]
\begin{center}
\includegraphics[width=0.7\linewidth]{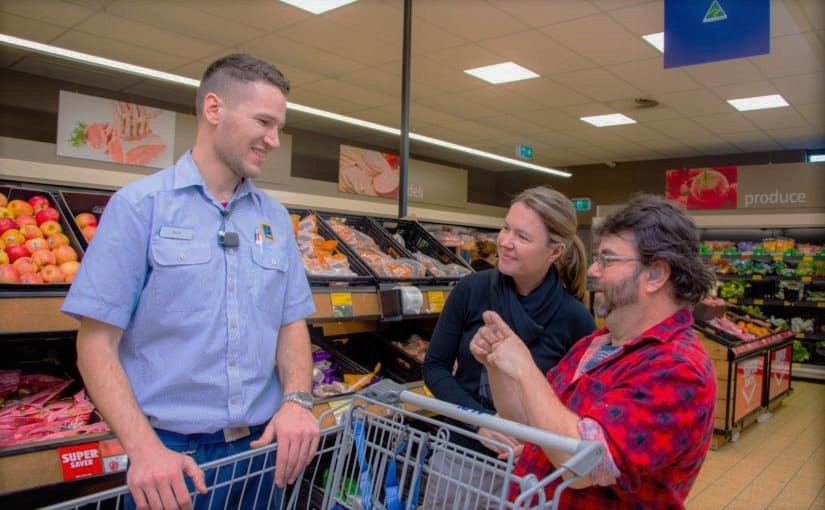}   
\end{center}
   \caption{A scenario where a deaf person asks for help \cite{grocery}.}
\label{fig:long}
\label{Fig 00}
\end{figure}

Sign Language is the dominant form of communication language used in deaf society. Most of the people in the hearing and deaf communities are not familiar with sign language. To touch on the necessity of this language for both hearing and deaf people, let imagine you are in a grocery store. What will happen if a deaf person asks you to help him/her? (See Figure \ref{Fig 00}). This is a challenging situation. If you do not know sign language, it would be useful to be able to use an application to translate a spoken language into sign language and vise versa. This example is just one situation among many others in which being familiar with sign language is useful. Actually, developing efficient bidirectional sign language translation systems requires expertise in a wide range of fields, including Computer Vision (CV), Computer Graphics (CG), Natural Language Processing (NLP), Human-Computer Interaction (HCI), Linguistics, and Deaf culture. While the vast majority of communication technologies have been developed for spoken/written language, sing languages have been excluded in most of these technologies. Furthermore, most hearing people do not know sign language. The results will be the existence of many communication barriers for deaf people in society. According to the World Health Organization (WHO) report in 2020, there are more than 466 million deaf people in the world \cite{who1} using different forms of sign languages, such as American Sign Language (ASL) \cite{asl}, Argentine Sign Language, \cite{argentina}, Polish Sign Language \cite{poland}, German Sign Language \cite{Germany}, Greek Sign Language \cite{Greek}, Spanish Sign Language \cite{spain}, Chinese Sign Language \cite{china}, Korean Sign Language \cite{Korean}, Persian Sign Language \cite{Iran}, just to mention a few. This report shows the necessity of study, research, and technology development in this area.

In general terms, the translation task is an important component in recent technologies developed by multinational institutions to aid in internal and external communications \cite{Bragg}. In more detail, there are two specific translation tasks in a bidirectional sign language translation system: Sign Language Recognition (SLR) and Sign Language Production (SLP). The former is defined as a translation task from the sign language into the spoken language. However, this task can be configured as a vision-based recognition task. The latter translates the spoken language into sign language. Both SLR and SLP tasks are fundamental to make a bidirectional translation system applicable in real-life applications. Compared to other languages, the translation task is more challenging in sign language. Some factors that contribute to such complexity are unfamiliarity of people with sign language, complex patterns in different signs, lack of a specific standard for sign languages, and the challenges corresponding to the vision-based tasks. Furthermore, the visual variability of signs is challenging, as it is affected by hand-shape, palm orientation, movement, location, facial expressions, and other non-hand signals. These differences in sign appearance produce a large intra-class variability and low inter-class variability. This makes it hard to provide a robust and universal system capable of recognizing different sign types. Another challenge is developing a photo-realistic SLP system to generate the corresponding sign digit/word/sentence from a text/voice in spoken language in a real-world situation. The challenge corresponding to the grammatical rules and linguistic structures of sign language needs to be considered. Translation between spoken and sign language is a complex problem. This is not a simple mapping problem from text/voice to signs word-by-word. This challenge comes from the differences between the tokenization and ordering of words in the spoken and sign languages. Another complex condition is related to the application area. Most of the applications in sign language focus on sign language recognition in different areas, such as robotics \cite{Dawes}, human–computer interaction \cite{Bachmann}, education \cite{Darabkh}, computer games \cite{Roccetti}, recognition of children with autism \cite{cai}, automatic sign-language interpretation \cite{yang}, decision support for medical diagnosis of motor skills disorders \cite{Butt}, home-based rehabilitation \cite{Cohen} \cite{Morando}, and virtual reality \cite{Vaitkev}. This is due to the misunderstanding of hearing people. Most of them think that deaf people are much more comfortable with reading spoken language; therefore, it is not necessary to translate the reading spoken language into sign language. This is not true since there is no guarantee that a deaf person is familiar with the reading and writing forms of a spoken language. Generally, these two forms of language are completely different from each other. Most of the sign languages, do not have a standard written form.

Nevertheless, despite the aforementioned challenges, some methods have been proposed with different degrees of success in both SLR and SLP \cite{RastgooSurvey}\cite{RastgooSLP}. While SLR has rapidly advanced in recent years \cite{Rastgoo-rbm}\cite{Rastgoo-multiview}\cite{Rastgoo-video}\cite{Rastgoo-pose-aware}\cite{Rastgoo-svd}\cite{RastgooSurvey}, SLP is still a very challenging problem, involving an interpretation between visual and linguistic information \cite{stoll2020}. Proposed systems in SLR generally map signs into the spoken language in the form of text transcription or speech \cite{RastgooSurvey}. However, SLP systems perform the reverse procedure. There are some accurate and well-detailed surveys in the SLR \cite{Ghanem}\cite{RastgooSurvey} but only one detailed discussion has been presented in SLP \cite{RastgooSLP}. Here, we extend our previous survey \cite{RastgooSLP}, and review recent advances in SLP and related areas using deep learning. To have more realistic perspectives to sign language, we present an introduction to the Deaf culture, Deaf centers, psychological perspective of sign language, the main differences between spoken language and sign language. Furthermore, we present the fundamental components of a bi-directional sign language translation system, discussing the main challenges in this area. Also, the backbone architectures and methods in SLP are briefly introduced and the proposed taxonomy on SLP is presented. Finally, a general framework for SLP and performance evaluation, and also a discussion on the recent developments, advantages, and limitations in SLP, commenting on possible lines for future research are presented.
The remainder of this paper is organized as follows. To have more realistic perspectives to sign language, we present an introduction to the Deaf culture in section 2. Some of the most familiar Deaf centers are presented in section 3. After that, we briefly discuss the psychological perspective of sign language in section 4. The main differences between spoken language and sign language are reviewed in section 5. section 6 presents the fundamental components of a bi-directional sign language translation system. In section 7, we dive into the SLP problem, discussing the main challenges in this area. The backbone architectures and methods in SLP are briefly introduced in section 8. Our taxonomy on SLP is presented in section 9. A general framework for SLP and performance evaluation, and also a discussion on the recent
developments, advantages, and limitations in SLP, commenting on possible lines for future research are presented in section 10, 11, and 12, respectively.

\section{Deaf culture}
Deaf Culture is the heart of the Deaf community. Language and culture are intertwined and inseparable passed down through generations of Deaf people. The Deaf community is not based on geographic vicinity. Generally, it contains two groups: the culturally Deaf people and the other individuals who use sign language. Actually, the Deaf community brings together these two groups. Furthermore, the intuition behind the Deaf culture is that it helps the Deaf people who are educated at residential Deaf schools to develop their own Deaf network once they graduate and keep in touch with everyone. Most of these Deaf people take on leadership positions in the Deaf community, organize Deaf sports, community events, etc, and become the core of the Deaf community. A key point is that this community needs to be sure that their language and heritage are passed to other peers and the next generation. They also form connections with parents and siblings of Deaf children to extend the community circle for Deaf children.

Suppression of sign language communication is a sample of cruelty against the Deaf community. One example is dated back to 1889 where an international congress of largely hearing educators of deaf students announced that sign language should be substituted with spoken language \cite{Lane}. Afterward, oralism, as a teaching system for Deaf people to communicate using speech and lip-reading instead of sign language, was extensively enforced. Since then, Deaf communities have struggled to use sign languages in schools, work, and public life \cite{Geers}. Furthermore, linguistic advancements have assisted sign languages so that they can be used as natural languages \cite{Stokoe}. Also, the role of the legislation can not be ignored in helping to establish legal support for sign language education and usage \cite{UN}. Considering this historical struggle can help the researchers to have a sense of the necessity of the translation and recognition systems for sign languages applicable in the real-life of the Deaf community \cite{Bragg}. 

\section{Deafness centers}
Since we maybe refer to the reports of the deafness centers in this survey, here, we present some most-used deafness centers to access data of Deaf community, just mentioned a few:

\begin{itemize}
    \item World Health Organization (WHO) \cite{WHO},
    \item National Institute on Deafness and Other Communication Disorders (NIDCD) \cite{NIDCD},
    \item Centers of Disease Control and Prevention \cite{CDC},
    \item National Deaf Center (NDC) \cite{NDC},
    \item Hearing, Speech and Deaf Center (HSDC) \cite{HSDC},
    \item Center for Hearing and Deaf Services (HDS) \cite{HDS},
    \item Deaf and Hard of Hearing Program \cite{DHHP},
    \item Manchester Centre for Audiology and Deafness (ManCAD) \cite{Manchester},
    \item Northern Virginia Resource Center for Deaf and Hard of Hearing Persons \cite{Virginia},
    \item National Center on Deaf-Blindness (NCDB) \cite{National}.
\end{itemize}

These centers aim to provide the educational, clinical, and research services to the Deaf community.

\section{Psychological perspective of sign language}
As we stated before, developing an efficient bidirectional sign language translation system requires the study in a wide range of fields, including CV, CG, NLP, HCI, Linguistics, and Deaf culture. To this end, we present a brief discussion of the findings from developmental psychology, psycho-linguistics, cognitive psychology, and neuropsychological studies. Recent studies of attention and perception show that usage of sign language from an early age can boost some aspects of non-language visual perception, such as motion perception. Furthermore, neuropsychological and functional imaging studies indicate that left hemisphere regions are important in both sign and spoken language processing. The aphasia can be occurred in signers due to left hemisphere damage. Also, the existence of different modalities for language expression, such as oral–aural and manual–visual, makes room to explore different characteristics of human languages. 

From a pathological perspective, Deaf people have different degrees of hearing deviations from the standard/norm hearing level defined for hearing people. Generally, four levels of deafness are defined: mild, moderate Hearing, severe, and profound hearing loss. This perspective is traditionally acceptable by a majority of non-deaf professionals who interact with the Deaf Community only on a professional basis. So, this issue can be considered for hardware implementations of the proposed systems in SLR and SLP. Developing such a system can make a room for the Deaf community to overcome the communication barriers and help them to keep motivated.

\section{Sign languages vs. spoken languages}
Generally, there are two main types of languages used in the community: spoken and sign. While these two language types are different from each other, both of them should be viewed as natural languages. The main difference between them refers to the way that they convey information. The spoken language is understood as an auditory/vocal language. It can also be considered as an oral language. The various sound patterns are used to convey a message. There are many linguistic elements in the spoken language, such as vowels, consonants, and tones. Making changes in these elements can lead to different meanings for the same set of words in the spoken language. In contrast to spoken languages, gestures and facial expressions play key roles to convey information in sign languages instead of vocal tracts. There are different sign languages in the world. Some of them are better-known, such as American Sign Language (ASL). In every country, there are one or more sign languages used by the Deaf community. While people think that sign languages have derived from spoken languages, they are independent of natural languages that have evolved over time. Sign language is a complex language that has specific linguistic properties. SLR is affected by the structural properties of sign language and occurs faster than spoken language recognition. While signs are articulated slower than spoken words, the proposition rate for sign and speech is identical. It should be noted that both languages can be used to convey all sorts of information, such as news, conversations about daily activities, stories, narrations, etc.

\section{Bi-directional sign language translation system}
\begin{figure}[h]
\begin{center}
\includegraphics[width=0.6\linewidth]{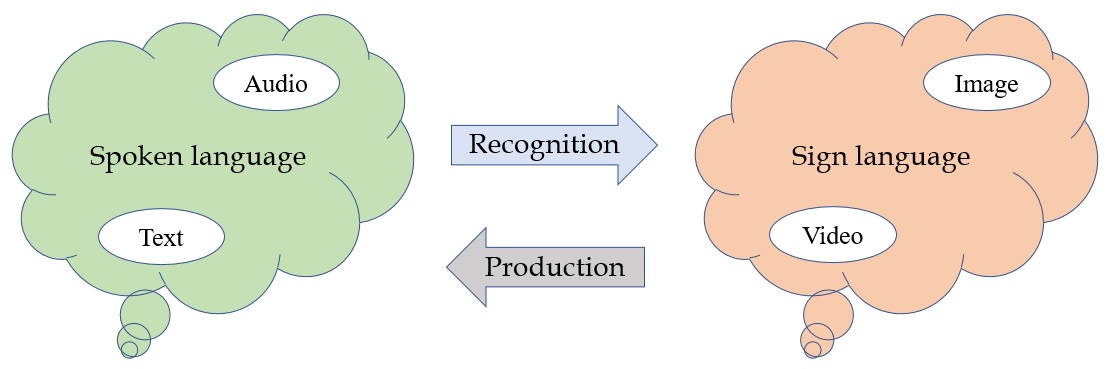}   
\end{center}
   \caption{ A bi-directional sign language translation system.}
\label{fig:long}
\label{Fig 0}
\end{figure}

As already discussed, to make a bi-directional sign language translation system, we need a system capable of translation from sign language into a spoken language (SLR) and vice versa (SLP) (See Fig. \ref{Fig 0}). While SLR has rapidly advanced in recent years, SLP is still a challenging problem. Since the details of SLR have been presented in some accurate and well-detailed surveys \cite{Ghanem} \cite{RastgooSurvey}, in this survey, we focus on SLP details of this bi-directional system and present more details of recent works in the SLP.

\section{SLP}
SLP is one of the main components of a bidirectional sign language translation system. This system can be used to facilitate easy and clear communication between the hearing and the Deaf communities. Furthermore, the necessity of such systems can also be considered as a psychological perspective for the Deaf community. In this section, we present more details on SLP.

\subsection{Problem definition}
The task of SLP can be defined as a video generation process from an input text. In more details, given a spoken language sentence, ${S}^{N} = \{{w}_{1},{w}_{2},...,{w}_{N}\}$, it is expected that the model generates a video with M frames, ${V}^{M} = \{{F}_{1},{F}_{2},...,{F}_{M}\}$, including a sign language video corresponding to the input sentence. Generally, there are some intermediate steps for the SLP task. During these steps, the input sentence from the spoken language is encoded into some representations to generate more accurate videos. We will review the proposed models and also the intermediate steps in SLP in this survey. 

\subsection{Challenges}
Here, we discuss the most important challenges in SLR.

\textbf{Interpretation between visual and linguistic information:}
SLP is still a very challenging problem involving an interpretation between visual and linguistic information \cite{stoll2020}. Proposed systems in SLR generally map signs into the spoken language in the form of text transcription \cite{RastgooSurvey}. However, SLP systems perform the reverse procedure. The challenges regarding mapping from the lingual domain into the visual domain still remain.

\textbf{Visual variability of signs:}
The visual variability of signs is one of the challenges in SLP, which is affected by hand-shape, palm orientation, movement, location, facial expressions, and other non-hand signals. These differences in sign appearance produce a large intra-class variability and low inter-class variability. This makes it hard to provide a robust and universal system.

\textbf{Photo-realistic SLP system:}
Another challenge is generating a photo-realistic sign video from a text or voice in spoken language in a real-world situation. This is important because it helps the generated videos to be truly understandable and accepted by Deaf communities. Thanks to the previous models based on graphical avatars and also recent neural SLP works that produce skeleton pose sequences, we need the systems that are understandable and acceptable to Deaf viewers \cite{Saunders-arxiv}.

\textbf{The grammatical rules and linguistic structures of the sign language:}
The challenge corresponding to the grammatical rules and linguistic structures of the sign language is another critical challenge in this area. Translation between spoken and sign language is a complex task. This is not a simple mapping problem from text/voice to sign word-by-word. Another issue worth mentioning here is the parallel nature of sign languages, how hands and face can operate simultaneously to convey lexical and grammatical information.

\textbf{Bilingual education:}
The brain has no preference for any type of languages. The only preference of the brain is that it expects to receive input from a complete and natural language. In this way, both spoken and sign languages can be used as inputs for the brain. Being bilingual, as a positive and desirable quality, the Deaf community can follow similar developmental paths as do monolinguals. This dual exposure can
lead to mental flexibility, creative thinking, and communication advantages \cite{Hamers}. Historically, a sign language has not been incorporated in the education of Deaf children \cite{Humphries} \cite{Grosjean} \cite{Swanwick}. Some earlier sign languages were not natural sign languages. They just used signing to deliver the content for the Deaf individuals who failed within an oral-only approach. The dissatisfaction with the educational outcomes of the Deaf community led to a bilingual design that placed sign language at the same level as the spoken/written language. To develop functional bilingual systems, the full development of two languages is crucial. In such systems, the social and academic functions of both languages are considered and their consistent and strategic use is promoted in the environment. The final goal is to deliver content instruction in both languages making it a viable design for Deaf children \cite{Garate}.

\textbf{Application area:}
Millions of Deaf and hearing-impaired people live across the world. The predominant part of them lives in low-income and developing countries with low access to suitable ear and hearing care services. While hearing loss makes many difficulties in the corresponding community, many mainsprings of it can be prevented through public health measures. Rehabilitation, education, empowerment, and communication technology usage are some of the main solutions to solve the communication barriers for the hearing-impaired community and use the full potential of the Deaf and hearing-impaired people. To this end, we present compact information regarding the community affected by hearing loss to make a promising insight and humanitarian motivation in a research community. Considering the scope of this survey, this information can help to develop communication technologies compatible with the needs of the hearing-impaired community.\\
According to the World Health Organization (WHO) report, 1.5 billion people live with some degrees of hearing loss. It is predicted that by 2050 approximately 2.5 billion people will have some degree of hearing loss \cite{report}. Some critical points need to be considered for application development in this area:
\begin{enumerate}
    \item \textbf{Geography:} 80 \%\ of people with hearing loss live in low-income countries. These people cannot easily access assistive technologies to improve their communication quality.
    
    \item \textbf{Age:} Another challenge is the hearing loss outbreak with age. Age is an important predictor of hearing loss among adults aged 20-69, with the maximum amount of hearing loss in the 60 to 69 age group. Nearly 25 \%\ of people older than 60 years are affected by hearing loss. This challenge can be considered for adopting the assistive technologies with the special physical and mental situations of people older than 60 years. Nearly 15 \%\ of American adults (37.5 million) aged 18 faces hearing loss. Furthermore, about 2 to 3 out of every 1000 American children are affected by hearing loss. Considering the difference of hearing loss definition in different ages, the age factor needs to be considered for application development. Generally, a hearing loss greater than 40 (dB) and 30 dB is defined for adults and children, respectively. Due to this difference and the other communication requirements corresponding to different age groups, the age factor is an important factor for application development in the field.
    
    \item \textbf{Gender:} According to the reports of WHO, men are approximately twice as likely as women to have hearing loss among adults aged 20-69. This is due to this fact that men are usually work in louder environments.
    
    \item \textbf{Sign language:} As we discussed in the previous sections, most Deaf people are not familiar with sign language. This makes it hard to develop communication tools for sign language translation.
\end{enumerate}

Considering all of these critical points, developing effective applications is challenging. Although different applications have been developed in recent years \cite{Zhao}\cite{Smith}\cite{Zij}\cite{Huenerfauth}\cite{Huenerfauth2004}\cite{Kanis}\cite{Hanke} \cite{Bangham}\cite{Veale}\cite{Karpouzis}\cite{Dangsaart}\cite{Jemni}\cite{SignSynth}, more endeavor is necessary to develop real-time applications for bi-directional translation from sign language to spoken language and vise versa. Furthermore, most of the applications in sign language focus on the recognition task, such as robotics \cite{Dawes}, human–computer interaction \cite{Bachmann}, education \cite{Darabkh}, computer games \cite{Roccetti}, recognition of children with autism \cite{cai}, automatic sign-language interpretation \cite{yang}, decision support for medical diagnosis of motor skills disorders \cite{Butt}, home-based rehabilitation \cite{Cohen} \cite{Morando}, and virtual reality \cite{Vaitkev}. This is due to a common misunderstanding by hearing people that the Deaf people are much more comfortable with reading spoken language; therefore, it is not necessary to translate the reading spoken language into sign language. This is not true since there is no guarantee that a deaf person is familiar with the reading/writing forms of a spoken language. In some languages, these two forms are completely different from each other. 

\textbf{Real-time communication:}
For now, accessibility in SL is mainly achieved by pre-recorded videos. This cannot enable real-time interaction for the content provider. To have an automatic sign recognition system applicable in a mutual interaction between a hearing-impaired/Deaf user and a hearing user or a digital assistant in real-time, we need low-complex and fast models. Using such models, the Deaf community can simply communicate with other people in different locations, such as schools, banks, hospitals, trains, University, just to mention a few. A translation system could be vision-based or sensor-based, depending on the type of input it receives. To date, most of the current commercial systems for sign language translation are sensor-based, which are expensive and not user-friendly. Vision-based sign translation systems are necessary but should overcome many challenges to build a system applicable in real-time communication.

\textbf{Sign anonymization:}
The purpose of sign anonymization is to ensure that no personal information of the signers is shared with the community. Furthermore, providing realistic, human-like, and anonymized animations would ensure higher acceptability and comprehensibility than actual signing avatars. In sign language, complete anonymization of the video data is not possible because both the face and hands of the signers must be fully visible so that the content can be understandable. Since most Deaf people have challenges in communication through written content, the need for producing messages anonymously is an important demand of them. As a result, video is the main communication modality used by native signers. The development of virtual signers is thus expanding, in order to make written material on the internet more available to deaf users.

\section{Backbone architectures and methods}
In this section, we review the most-used architectures and methods in SLP: Convolutional Neural Networks (CNNs), Recurrent Neural Networks (RNNs), generative models, motion capture, and signing avatars.

\subsection{CNNs}
One of the basic deep learning-based building blocks designed for visual reasoning is convolutional layers. Using these layers, CNNs effectively model the spatial structure of images \cite{Lecun}. In SLP, CNNs are the foundation of the proposed models. However, the CNNs performance faces some challenges. One challenge in CNNs is corresponding to the limited receptive field, introduced as a kernel size. As some solutions to this challenge, we can take into account: stacking more convolutional layers \cite{Jain}, increasing the kernel size, linearly fusing multiple scales \cite{Mathieu}\cite{Denton}, using dilated convolutions to include long-range spatial dependencies \cite{Yu}, extending the receptive fields \cite{Chen}\cite{Luo}, sub-sampling, or using residual connections \cite{He}\cite{Villegas}. Another challenge in CNNs is the lack of temporal learning corresponding to the image sequences. To properly address this challenge, 3D convolutions are used as a promising alternative to recurrent modeling. Several models have been proposed to sign language using 3D convolutions \cite{Sharma}\cite{hammadi}\cite{Sripairojthikoon}\cite{Rastgoo-multiview}. However, the 3DCNN models are not generally as powerful as the sequence learning models such as RNN, LSTM, and GRU.  

\subsection{Transformer}
The main intuition behind recurrent models is modeling the temporal representation of sequential data, such as image sequences. Deep recurrent networks demonstrated great success in different sequence learning tasks, such as machine translation \cite{Siddique}, speech recognition \cite{Graves}, video captioning \cite{Pei}, video prediction \cite{Villegas}, SLR \cite{Rastgoo-multiview}, and SLP \cite{RastgooSLP}. However, there are some limitations in these networks, such as vanishing and exploding gradient. To mitigate these challenges, the classical RNNs were extended to more sophisticated recurrent models, such as Long Short-Term Memory (LSTM) \cite{LSTM} and Gated Recurrent Unit (GRU) \cite{Cho}. Different works have explored different modifications of the extended recurrent models, such as applying the LSTM-based models to the image space \cite{Shi}, using multidimensional LSTM (MD-LSTM) \cite{Graves2007}, using the stacked recurrent layers to include abstract spatio-temporal correlations \cite{Finn}\cite{Lotter}, and addressing the duplicated recurrent representations \cite{Zhan}. In addition, the Transformer models have recently gained results due to using the self-attention mechanism and parallel computing. In most of the models for SLP, a recurrent model is used for the temporal representation of sequential data.

\subsection{Generative models}
Generative modeling is an unsupervised learning task in machine learning. It involves automatically discovering and learning the regularities or patterns in input data. Such a model can generate or output plausible examples. Generally, there are two main categories for model learning: discriminative and generative. While a discriminative model learns the decision boundaries between the classes, a generative model learns the real distribution of each class. In other words, a generative model learns the joint probability distribution p(x,y) to predict the conditional probability using the Bayes Theorem. From the other side, a discriminative model learns the conditional probability distribution p(y|x). Both of these models generally fall into the supervised learning problems. The goal of generative models is to generate new samples from the same distribution, given some training data. In the learning procedure, the distribution of the real and generated data gets closer to each other. This is done by explicitly, e.g VAEs, or implicitly, e.g. GANs, estimating a density function from the real data. In SLP, generative models are used to generate more realistic and plausible videos, considering sign language challenges.

\subsection{Motion capture and signing avatars}
Motion capture, mocap for short, is defined as the process of recording the movement of objects/people. Different application areas use mocap for their requirements fulfilment, such as sports \cite{sport}, entertainment \cite{entertainment}, gaming industry \cite{animation}, robotics \cite{robotics}, automotive \cite{automotive}, and construction \cite{reconstruction}. In movie production and video game development, mocap refers to recording actions of human actors and utilizing that information to animate digital character models in a 2D/3D computer animation. During the mocap sessions, the movements of one or more actors are sampled many times per second. The mocap aims to record only the movements of the actors, not their visual appearance. Finally, the animation data is mapped to a 3D model in a way that the model performs the same actions as the actor.\\  
The mocap process has several advantages over traditional computer animation of a 3D model, such as lower latency for data recording, the ability to produce a large amount of data within a given time, the ability to create complex movement and realistic physical interactions. However, there are some challenges in mocap usage. The need for special and expensive hardware/software to obtain and process the data, the need for specific requirements for the space that the mocap process is operated in, and the need for re-recording data instead of manipulating it in facing with problems are some of these challenges. \\
Signing avatars are an animated 3D model of the mocap data obtained using signers. Animating can be manually defined, captured from a human signer, or parametrically described. The signing avatars aim to assist the research community in making different applications more accessible to the Deaf community. Furthermore, they will also help address the lack of human interpreters. The goal is not to replace the human interpreters but rather to increase the amount of signed content available to the Deaf users. As another application of signing avatars, they can be used as assistive technologies for deaf students in school. Using these technologies, the interaction among Deaf and hearing students will be much more easier.\\ 
Recently, mocap data is used to edit and generate sign language samples. To this end, some motion edition operations, such as concatenation and mixing, are applied to mocap data to compose new utterances. This helps to facilitate the enrichment of the original mocap data, enhancing the natural look of the animation, and promoting the avatar’s acceptability. However, manipulating existing movements does not guarantee the semantic consistency of the reconstructed signs. Employing an expert user for constructing new utterances from linguistic patterns can be a primary solution to this challenge.

\section{SLP taxonomy}
\begin{figure}[t]
\begin{center}
\includegraphics[width=11cm,height=7cm,keepaspectratio]{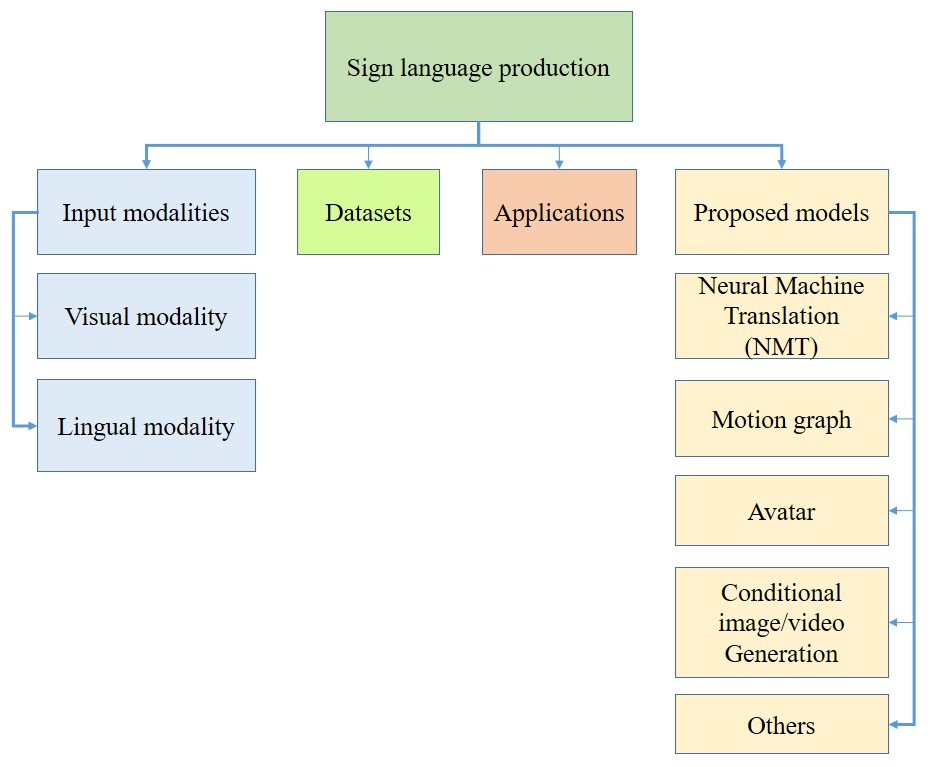}
\end{center}
   \caption{The proposed taxonomy of the reviewed works in SLP.}
\label{fig:long}
\label{Fig 1}
\end{figure}

In this section, we present a taxonomy that summarizes the main concepts related to deep learning in SLP. We categorize recent works in SLP providing separate discussions in each category. In the rest of this section, we explain different input modalities, datasets, applications, and proposed models. Figure \ref{Fig 1} shows the proposed taxonomy described in this section.

\subsection{Input modalities}

Generally, vision and language are two input modalities in SLP. While the visual modality includes the captured image/video data, the linguistic modality for the spoken language contains the text/audio input from the natural language. Computer vision and natural language processing techniques are necessary to process these input modalities. While the visual modality is used in the training, the lingual modality is applicable in both the training and testing of the proposed models.

\textbf{Visual modality:} RGB and skeleton are two common types of input data used in SLP models. While RGB images/videos contain high-resolution content, skeleton inputs decrease the input dimension necessary to feed to the model and assist in making a low-complex and fast model. The spatial features corresponding to the input image can be extracted using computer vision-based techniques, especially deep learning-based models. In recent years, Convolutional Neural Networks (CNN) achieved outstanding performance for spatial feature extraction from an input image \cite{Rastgoo-cnn}. Furthermore, generative models, such as Generative Adversarial Networks (GAN), can use CNN as an encoder or decoder block to generate a sign image/video. Due to the temporal dimension of RGB video inputs, the processing of this input modality is more complicated than the RGB image input. Most of the proposed models in SLP use the RGB video as input \cite{Camgoz,SaundersBMVC,Saunders,stoll2020}. An RGB sign video can correspond to one sign word or some concatenated sign words, in the form of a sign sentence. GAN and LSTM are the most used deep learning-based models in SLP for static and temporal learning in the visual input modalities. While successful results have been achieved using these models, more effort is necessary to generate more lifelike sign images/videos in order to improve the communication interface with the Deaf community.

\textbf{Lingual modality:} Text input is the most common form of linguistic modality. To process the input text, different models are used \cite{See,Sutskever}. 
Among the deep learning-based models, the Neural Machine Translation (NMT) model is the most used model for input text processing. The Seq2Seq models \cite{Sutskever}, such as Recurrent Neural Network (RNN)-based models, proved their effectiveness in many tasks. While successful results were achieved using these models, more effort is necessary to overcome the existing challenges in the translation task. One challenge in translation task is related to domain adaptation due to different words styles, translations, and meanings in different languages. Thus, a critical requirement of developing machine translation systems is to target a specific domain. Transfer learning, training the translation system in a general domain followed by fine-tuning on in-domain data for a few epochs is a common approach in coping with this challenge. Another challenge is regarding the amount of training data. Since the main property of deep learning-based models is the mutual relation between the amount of data and model performance, a large amount of data is necessary to provide a good generalization capability in the model. Another challenge is the poor performance of machine translation systems on uncommon and unseen words. To cope with these words, byte-pair encoding, such as stemming or compound-splitting, can be used for rare words translation. As another challenge, the machine translation systems are not properly able to translate long sentences. However, the attention model \cite{Vaswani} partially deals with this challenge for short sentences. Furthermore, the challenge regarding the word alignment is more critical in the reverse translation, that is translating back from the target language to the source language.

\subsection{Datasets}
While there are some large-scale and annotated datasets available for sign language recognition \cite{RastgooSurvey}, there are only a few publicly available large-scale datasets for SLP. Two public datasets, RWTH-Phoenix-2014T \cite{Necati2018} and How2Sign \cite{Duarte} are the most used datasets in sign language translation. The former includes German sign language sentences that can be used for text-to-sign language translation. This dataset is an extended version of the continuous sign language recognition dataset, PHOENIX-2014 \cite{Forster}. RWTH-PHOENIX-Weather 2014T includes a total of 8257 sequences performed by 9 signers. There are 1066 sign glosses and 2887 spoken language vocabularies in this dataset. Furthermore, the gloss annotations corresponding to the spoken language sentences have been included in the dataset. The later dataset, How2Sing, is a recently released multi-modal dataset used for speech-to-sign language translation. This dataset contains a total of 38611 sequences and 4k vocabularies performed by 10 signers. Like the former dataset, the annotation for sign glosses has been included in this dataset.

\begin{figure}[t]
\begin{center}
\includegraphics[width=11cm,height=7cm,keepaspectratio]{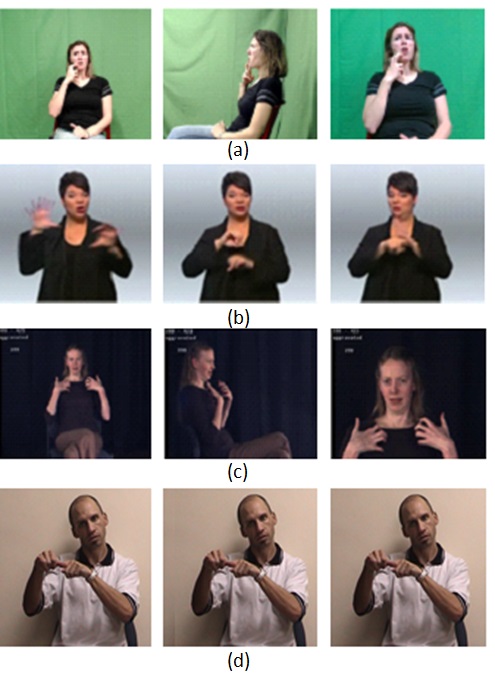}
\end{center}
   \caption{Samples of some most-used datasets: (a) How2Sign, (b) RWTH-PHOENIX-Weather 2014T, (c) ASLVID, (d) ATIS Corpus.}
\label{fig:long}
\label{Fig 2}
\end{figure}

\begin{figure}[t]
\begin{center}
\includegraphics[width=8cm,height=4cm,keepaspectratio]{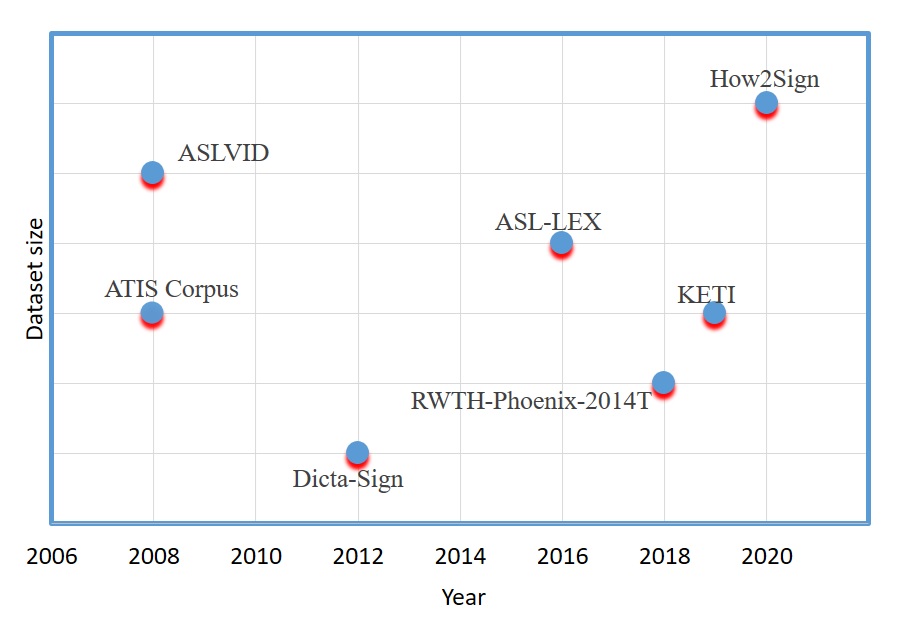}  
\end{center}
   \caption{SLP datasets in time. The number of samples for each dataset is shown in brackets}
\label{fig:long}
\label{Fig 3}
\end{figure}

Though RWTH-PHOENIX-Weather 2014T and How2Sign provided SLP evaluation benchmarks, they are not enough for the generalization of SLP models. Furthermore, these datasets just include German and American sentences. In line with the aim of providing an easy-to use application for mutual communication between the Deaf and hearing communities, new large-scale datasets with enough variety and diversity in different sign languages are required. The point is that the signs are generally dexterous and the signing procedure involves different channels, including arms, hands, body, gaze, and facial expressions simultaneously. Capturing such gestures requires a trade-off between capture cost, measurement (space and time) accuracy, and production spontaneity. Furthermore, different equipment is used for data recordings, such as wired Cybergloves, Polhemus magnetic sensors, headset equipped with an infrared camera, emitting diodes and reflectors. Synchronization between different channels captured by the aforementioned devices is key in data collection and annotation. Another challenge is related to the capturing complexity of the hand movement using some capturing devices, such as Cybergloves. Hard calibration and deviation during data recording are some difficulties of these acquisition devices. The synchronization of external devices, hand modeling accuracy, data loss, the noisy capturing process, facial expression processing, gaze direction, and data annotation are additional challenges. Given these challenges, providing a large and diverse dataset for SLP, including spoken language and sign language annotations, is difficult. Figure \ref{Fig 2} and Figure \ref{Fig 3} show some samples and also the timeline of existing datasets for SLP.

To make a sense of the existence spoken and sign language datasets, we review some datasets in machine translation for spoken to spoken language translation. Analyzing these datasets shows that there are more spoken datasets with more variety in the sample and language numbers, compared to sign language datasets. For example, the MUSE dataset includes bilingual dictionaries for 110 language pairs. For each language pair, the training and testing seed dictionaries include approximately 5000 and 1500 word pairs, respectively. Another dataset, namely OpenSubtitles, is a collection of multilingual parallel corpora obtained from a large database of movie and TV subtitles. OpenSubtitles contains a total of 1689 bitexts spanning 2.6 billion sentences across 60 languages. Multi30K is a multi-modal dataset obtained from the Flickr30k dataset. This dataset includes 31,014 images and the corresponding five English descriptions. The Flicker dataset includes 145,000 training, 5,070 validation, and 5,000 test descriptions. The Multi30K dataset aims to translate the Flicker30 descriptions to the German sentences. ASPEC dataset contains a Japanese-English paper abstract corpus of 3M parallel sentences (ASPEC-JE) and a Japanese-Chinese paper corpus of 680K parallel sentences. MLQA is a cross-lingual dataset containing over 5K Question Answering (QA) samples (12K in English) in SQuAD format in seven languages: English, Arabic, German, Spanish, Hindi, Vietnamese and Chinese. The MTNT dataset is a Machine Translation dataset that contains the noisy comments on Reddit and professionally sourced translation. The translation is between French, Japanese and French, with between 7k and 37k sentences per language pair. Table \ref{Table 1} summarizes the most-used datasets for SLP and also the datasets for spoken to spoken language translation.

\begin{table*}[h!]
\thispagestyle{empty}
\caption{\label{Table 1} SLP datasets in time.}
\begin{center}
{\small
 \noindent\begin{tabular}{p{1cm}p{3.7cm}p{2.5cm}p{1cm}p{4cm}p{1cm}p{1.3cm}}
 \hline
\textbf{Type} & \textbf{Dataset} & \textbf{Nationality} & \textbf{Level} & \textbf{Content type} & \textbf{Public} & \textbf{Year} \\
\hline\hline
& ASLLVD \cite{Athitsos}  & English (US) & Word & Video, Gloss, Trans.& Y & 2008\\
& ATIS Corpus \cite{Bungeroth}  & Multilingual & Sentence & Video, Gloss, Trans.& Y & 2008\\
& Dicta-Sign \cite{Matthes} & English (US) & Word & Video, Gloss, Trans.& Y & 2012\\
Sign & ASL-LEX \cite{Caselli} & English (US) & word & Video, Gloss, Trans.& Y & 2016\\
& RWTH-Phoenix-2014T \cite{Necati2018} & German & Sentence & Video, Gloss, Trans.& Y & 2018\\	
& KETI \cite{Ko} & Korean & Sentence & Video, Gloss, Trans.& N & 2019\\
& How2Sign \cite{Duarte} & English (US) & Sentence & Video, Gloss, Trans, Speech. & Y & 2021\\
\hline\hline
& OpenSubtitles \cite{OpenSubtitles} & Multilingual (60) & Sentence & Video, Trans. & Y & 2016\\
& Multi30K \cite{Multi30K} & English, German & Sentence & Image, Trans. & Y & 2016\\
Spoken & ASPEC \cite{ASPEC} & Japanese, English & Sentence & Text & Y & 2016\\
& MUSE \cite{MUSE1,MUSE2} & Multilingual (110) & Word & Text & Y & 2017\\
& MTNT \cite{MTNT} & Japanese, French & Sentence & Text & Y & 2018 \\
& MLQA \cite{MLQA} & Multilingual (7) & Sentence & Text & Y & 2019\\
\noalign{\smallskip}\hline\hline
\end{tabular}
 }
\end{center}
\end{table*}

\subsection{Applications and technologies}

\subsubsection{Applications}
With the advent of potent methodologies and techniques in recent years, machine translation applications have become more efficient and trustworthy. One of the early efforts on machine translation is dated back to the sixties, where a model was proposed to translate from Russian to English. This model defined the machine translation task as a phase of encryption and decryption. Nowadays, the standard machine translation models fall into three main categories: rule-based grammatical models, statistical models, and example-based models. Deep learning-based models, such as Seq2Seq and NMT models, fall into the third category, and showed promising results in SLP.

To translate from a source language to a target language, a corpus is needed to perform some preprocessing steps, such as boundary detection, word tokenization, and chunking. While there are different corpora for most spoken languages, sign language lacks from such a large and diverse corpus. Since Deaf people may not be able to read or write in spoken language, they need some tools for communication with other people in society. Furthermore, many interesting and useful applications on the Internet are not accessible for the Deaf community. However, we are still far from having applications accessible for Deaf people with large vocabularies or sentences from real-world scenarios. One of the main challenges for these applications is a license right for usage. Only some of these applications are freely available. Another challenge is the lack of generalization of current applications, which are developed for the requirements of very specific application scenarios.

Here, we present some of the most used projects in sign language translation.

\textbf{Translation from English to ASL by Machine (TEAM) project:} An English translation system uses the grammar rules to create an American Sign Language (ASL) syntactic structure. Using the signing avatar, this project achieved successful performance for generating aspectual and adverbial information in ASL \cite{Zhao}.

\textbf{Machine Translation of Weather reports from English to ASL project:} Using the freely available Perl modules, some packages are designed to employ ASL grammar rules and generate the fluent ASL words \cite{Smith}.

\textbf{South African Sign Language Machine Translation (SASL-MT) project:} Like the TEAM Project \cite{Zhao}, SASL-MT uses the rule-based transfer mechanism from English to ASL. The SASL-MT is freely available for the Deaf community in specific domains, such as clinics, hospitals, and police stations. While this project is still under development, no evaluation results have been reported \cite{Zij}.

\textbf{Multi-path architecture for Sign Language Machine Translation (SLMT): } Using the virtual reality scene, a multi-channel architecture is proposed to include supplementary information of ASL. This project aims to generate spatially complex ASL words \cite{Huenerfauth}\cite{Huenerfauth2004}.

\textbf{Czech Sign Language Machine Translation: } Using the computer animation techniques, hand articulations are generated using an automatic process. Translation from spoken Czech to Signed Czech is a primary goal of this project. More than 3000 simple or linked sign vocabularies of Czech sign language are included in the dictionary of this project, which is a successful improvement in Czech sign language translation \cite{Kanis}\cite{Hanke}.

\textbf{Virtual signing, capture, animation, storage and transmission (ViSiCAST) Translator: } This project is proposed to translate from English text to British Sign Language (BSL). Using the grammar rules and symbolic representation, natural movements in the sing words are modeled. This project has successfully developed an avatar-based signing system for BSL \cite{Bangham}.

\textbf{ZARDOZ System: } This system is an English translation system using artificial intelligence knowledge representation, metaphorical reasoning, and blackboard system architecture. The main advantage of this system is the efficient performance for processing semantic information. The contributors of this project aim to improve this system using intelligent linguistic technologies \cite{Veale}.

\textbf{Environment for Greek Sign Language Synthesis: } This system includes an educational platform for deaf children. Virtual character animation techniques are used for sign sequence synthesis and lexicon-grammatical processing of Greek sign language sequences \cite{Karpouzis}.

\textbf{Thai-Thai Sign Machine Translation (TTSMT): } This model is a multi-phase approach to translate the Thai text into Thai Sign language. This system has been developed using the spatial grammatical order of the sign words and evaluated on the frequently used sign words in daily communication \cite{Dangsaart}.

\textbf{Web-Based Interpreter of Sign Language (WebSign): } WebSign is a project to develop a web-based tool for information processing. It includes a Plug-in to play the ASL sign. WebSign, as an avatar-based technology, generates a real-time and online interpretation in sign language that decreases the communication barriers between Deaf and hearing people \cite{Jemni}.

\begin{figure*}[t]
\begin{center}
\includegraphics[width=0.9\linewidth]{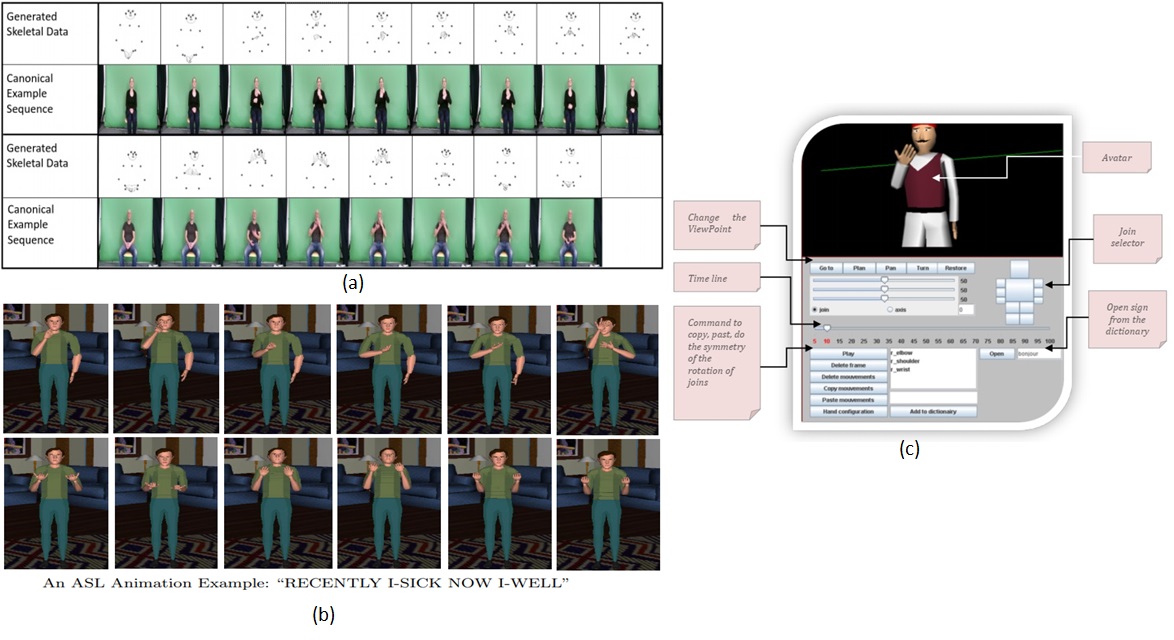}   
\end{center}
   \caption{A glance at some projects in SLP: (a) SignSynth \cite{SignSynth}, (b) TEAM Project \cite{Zhao}, (c) WebSign \cite{Jemni}.}
\label{Fig 4}
\end{figure*}

\textbf{Sign Language Synthesis Application (SignSynth): } SignSynth is a Deep Learning-based project for video generation from the human pose sequences using a generative model. One of the main advantages of this project is the capability of producing natural-looking sign videos using a fully automatic approach \cite{SignSynth}. Fig. \ref{Fig 4} shows a summary view of some of these projects.


\subsubsection{Assistive technologies for SLP}
In this section, we review the exiting assistive technologies for Deaf and hearing-impaired people to get an insight into the researchers in SLP and also make a bridge between them and the corresponding technology requirements. Considering the pros and cons corresponding to the exiting assistive technologies, researchers in SLP can develop much more real-world technologies for Deaf people. These technologies fall into three device categories: hearing technology, alerting devices, and communication support technology. Here, we present a quick recap on each category and discuss the promises and challenges.

\textbf{Hearing technology: }
Hearing technology contains devices employed to enhance the sound level available to a listener. So, it is not suitable for deaf people with a complete loss of their hearing ability. Main devices used in this technology include Hearing Aids (HAS) devices, assistive listening devices, Personal Sound Amplification Products (PSAPs), and cochlear implants \cite{Marion2003}. Here, we present a quick definition of each device. Furthermore, Fig \ref{Fig 4} shows some graphical views of these devices.

\textbf{HAs devices:}  HAs devices are sound-amplifying devices employed to enhance the hearing quality for impaired-hearing people \cite{EHIMA}.

\textbf{Assistive listening devices (ALD):} Like HAS, ALD are used to amplify the sounds, especially in noisy backgrounds \cite{NAD}.

\textbf{PSAPs:} PSAPs are devices that increase the sound levels and reduce the background noise.

\textbf{Cochlear Implant (CI):} CI, considered as an artificial cochlea, is a surgically-implanted sensor for sound to electrical waves conversion.
 
\textbf{Alerting devices: }
Since hearing ability is not obligatory in the alerting or alarm systems, these systems can be used in the Deaf community. Alert systems usually use light, vibrations, or a combination of them to make an alert notification for users. There are various types of alerting devices, including clocks and wake-up alarm systems, household device alerts, doorbell and telephone alerts, and baby monitoring devices. These devices employ the remote receivers placed around the house \cite{Lucker}.

\textbf{Communication support technology: }
Communication support technologies aim to facilitate communication between different communities. They generally fall into two main categories: telecommunication services and person-to-person interactions. The former category contains a variety of standard technologies, such as physical and virtual keyboards, touch screens, video calling, captioning for phone calls, voice to sign language translation, recognition text messaging, and text-based technology (such as WhatsApp \cite{whatsapp}, FB Messenger \cite{Facebook}, and Snapchat \cite{Snapchat}). The latter category covers person-to-person interactions using picture boards, keyboards, touch screens, display panels, and speech-generating devices. Table \ref{Table 2 } shows a statistical report on the existing technology products based on the EASTIN database \cite{EASTIN}.

\begin{table}
\caption{\label{Table 2} Statistical report on the existing technology products based on the EASTIN database \cite{EASTIN}.}
\begin{center}
\begin{tabular}{p{5cm}p{2cm}}
\hline
\textbf{Device type} & \textbf{Product Number}\\
\hline\hline
Hearing technology & 300\\
Alerting devices & 173\\
Communication support technology & 223\\
\hline
\end{tabular}
\end{center}
\end{table}

\textbf{Promises and challenges: }
Actually, research on Deaf and hearing-impaired technologies coincides with research on mainstream technology. While Deaf and hearing-impaired people contribute to mainstream technology \cite{WFD}, most of the communication technologies only support spoken or written languages, excluding sign language. They are among the first adopters of recent technologies such as Skype \cite{skype}, Google Hangout \cite{Google-Hangout}, FaceTime \cite{FaceTime}, Instant Messaging (IM) \cite{WFD}, text-to-speech and speech recognition software \cite{siri}\cite{Cortana}, WhatsApp \cite{whatsapp}, and Imo \cite{Imo}. While these tools have become an important part of our life, Deaf and hearing-impaired people have many problems with using these technologies. Recent developments in sign language recognition and production aim to facilitate bidirectional communication between Deaf and hearing people. The sensory substitution across sensory systems, such as vibratory and visual-auditory substitutions, is an active research area aiming to make a real-life perceptual experience of hearing for Deaf and hearing-impaired people. For example, one can imagine a technology that assists a Deaf person go through a musical experience translated into another sensory modality. The recent advances in computer vision algorithms, especially deep learning models, made room to develop some applications in sign language. As we presented in the previous sections, the recent advances in SLP are promising. However, more endeavor is indispensable to provide a fast processing model in an uncontrolled environment considering rapid hand motions. It is clear that technology standardization and full interoperability among devices and platforms are prerequisites to having real-life communication between two communities.

\subsection{Proposed models}

In this section, we review recent works in SLP. These works are presented and discussed in five categories: Avatar approaches, NMT approaches, Motion Graph (MG) approaches, Conditional image/video Generation approaches, and other approaches. Table \ref{Table 3} and Table \ref{Table 4} present a summary of the main characteristics and details of the reviewed models. 

\subsubsection{Avatar Approaches}

In order to reduce the communication barriers between hearing and hearing-impaired people, sign language interpreters are used as an effective yet costly solution. To inform deaf people quickly in cases where there is no interpreter on hand, researchers are working on novel approaches to providing the content. One of these approaches is sign avatars. Avatar is a technique to display the signed conversation in the absence of the videos corresponding to a human signer. To this end, 3D animated models are employed, which can be stored more efficiently compared to videos. The movements of the fingers, hands, facial gestures, and body can be generated using the avatar. This technique can be programmed to be used in different sign languages. With the advent of computer graphics in recent years, computers and smartphones can generate high-quality animations with smooth transitions between the signs. To capture the motion data of deaf people, some special cameras and sensors are used. Furthermore, a computing method is used to transfer the body movements into the sign avatar \cite{Kipp-Avatar}.

Two ways to derive the sign avatars include the motion capture data and parametrized glosses. In recent years, some works have been developed exploring avatars animated from the parametrized glosses. VisiCast \cite{Bangham}, Tessa \cite{Cox}, eSign \cite{Zwitserlood}, dicta-sign \cite{Efthimiou}, JASigning \cite{VHG}, and WebSign \cite{Jemni} are some of them. These works need the sign video annotated via the transcription language, such as HamNoSys\cite{Prillwitz} or SigML \cite{Kennaway}. However, under-articulated, unnatural movements, and missing non-manuals information, such as eye gaze and facial expressions, are some challenges of the avatar approaches. These challenges lead to misunderstanding the final sign language sequences. Furthermore, due to the uncanny valley, the users do not feel comfortable \cite{Mori} with the robotic motion of the avatars. To tackle these problems, recent works focus on the annotation of non-manual information such as face, body, and facial expression \cite{Ebling-2015,EblingJohn-2013}. For instance, Kipp et al. \cite{Kipp-Avatar} proposed two techniques, the torso and the noise methods, to aid manual animation and supplement procedural generated avatar movements systems such as \cite{Hanke}\cite{Delorme}. The first technique is an extension to any limb system and helps to automatically rotate the torso and spine of an avatar. This rotation supports the specified arm motions from the linguistic model. The second technique generates the motion in the held joints. Evaluation results show the effectiveness of these techniques. Though, the accurate alignment and articulation of this information are challenging \cite{McDonald}\cite{Kipp-Avatar}. More concretely, three steps have been included in the proposed model by \cite{McDonald}: movement of the spine, spreading the effect over the spine, and shoulder movement. To this end, the following parameters are included in the model:
\begin{equation}
    V_{R} = A_{R} - S_{R},
\end{equation}
\begin{equation}
    V_{L} = A_{L} - S_{L},
\end{equation}
\begin{equation}
    V_{reach} = \frac{V_{R} + V_{L}}{2},
\end{equation}
where $V_{R}, V_{L}, and V_{reach}$ are the displacement vectors from the right and left shoulders and also the average of these two vectors, respectively. To compute both the bend angle and direction, the torso must be rotated in the direction of $V_{reach}$. The user tests indicated that including such movements in the system would be highly beneficial. \\

Using the data collected from motion capture, avatars can be more usable and acceptable for reviewers (such as the Sign3D project by MocapLab \cite{Gibet}). Highly realistic results are achieved by avatars, but the results are restricted to a small set of phrases. This comes from the cost of the data collection and annotation. Furthermore, avatar data is not a scalable solution and needs expert knowledge to perform a sanity check on the generated data. To cope with these problems and improve performance, deep learning-based models, as the latest machine translation developments, are used. Generative models along with some graphical techniques, such as Motion Graph, are being recently employed \cite{stoll2020}.

\subsubsection{NMT approaches}
Machine translators are a practical methodology for translation from one language to another. The first translator comes back to the sixties where the Russian language was translated into English \cite{Hutchins}. The translation task requires preprocessing of the source language, including sentence boundary detection, word tokenization, and chunking. These preprocessing tasks are challenging, especially in sign language. Sign Language Translation (SLT) aims to produce/generate spoken language translations from sign language considering different word orders and grammar. The ordering and the number of glosses do not necessarily match the words of the spoken language sentences.

Nowadays, there are different types of machine translators, mainly based on grammatical rules, statistics, and examples \cite{Othman}. For instance, Othman and Jemni \cite{Othman} proposed a machine translation, namely IBM 1, by defining the translation probability for an English sentence $f = (f_{1}, f_{2}, ..., f_{N})$ of length $N$ to ans ASL sentence $e = (e_{1}, e_{2}, ..., e_{M})$ of length $M$ with an alignment of each ASL word $e_{j}$ to an English word $f_{i}$, considering the alignment function $a: j \to i$ as follows:\\

$p(e,a|f) = \frac{\epsilon}{(N+1)^{M}} \prod_{j=1}^{M}t(e_{j}|f_{a(j)})$,   (4)\\
    
where t is a conditional probability function. The alignment function $a$ maps each ASL output word $j$ to an English input position $a(j)$. The alignment probability distribution is also applied in this reverse direction. The combination of these two steps make the IBM 2 model, as follows:\\

$p(e,a|f) = {\epsilon} \prod_{j=1}^{M}t(e_{j}|f_{a(j)})a(a(j|j,N,M))$.         (5)\\

As an example-based methodology, some research works have been developed by focusing on translation from text into sign language using Artificial Neural Networks (ANNs), namely NMT \cite{Bahdanau}. NMT uses ANNs to predict the likelihood of a word sequence, typically modeling entire sentences in a single integrated model. Seq2seq model \cite{Sutskever}\cite{Cho2014}, as one of the most interesting breakthroughs in neural machine translations, consists of two Recurrent Neural Networks (RNNs). These RNNs form an encoder-decoder architecture to translate from a source sequence to a target sequence. This model aims to overcome the challenges in problems whose input and output sequences have different lengths with complicated and non-monotonic relationships. Considering the capabilities of the Long Short-Term Memory (LSTM) network \cite{LSTM-org} in learning the long-range temporal dependencies, the seq2seq model improved the translation performance of the model. More concretely, the LSTM aims to estimate the conditional probability $p(y_{1},...,y_{T'}|x_{1},...,x_{T})$, where $(x_{1},...,x_{T})$ and $(y_{1},...,y_{T'})$ are the input and output sequences, respectively. The length of input and output sequences may differ from each other. The LSTM network computes the conditional probability by first obtaining the fixed dimensional representation $v$ of the input sequence given by the last hidden state of the LSTM, and then computing the probability of the output sequence with a standard LSTM formulation. The initial hidden state is set to the representation $v$ of $x_{1},...,x_{T}$:

$p(y_{1},...,y_{T'}|x_{1},...,x_{T}) = \prod_{t=1}^{T^{'}}p(y_{t}|v,y_{1},...,y_{t-1})$.     (6)\\

where each $p(y_{t}|v,y_{1},...,y_{t-1})$ distribution is represented with a Softmax over all the words in the vocabulary. As a requirement, a special End-Of-Sentence symbol “<EOS>” is necessary to enable the model to define a distribution over sequences of all possible lengths. In addition to the LSTM Network, the Gated Recurrent Units (GRU) model \cite{Chung} can be used as an RNN cell. The seq2seq models proved their effectiveness in many sequence generation tasks by obtaining nearly human-level performance. However, there are some drawbacks to these models. One of them is corresponding to the fixed-size vector representation of the input sequences with different lengths. Vanishing gradient related to the long-term dependencies is another drawback of this model \cite{Ko}. To enhance the translation performance of long sequences, Bahdanau et al. \cite{Bahdanau} presented an effective attention mechanism. This mechanism was later improved by Luong et al. \cite{Luong}.

\begin{figure}[t]
\begin{center}
\includegraphics[width=9cm,height=5cm,keepaspectratio]{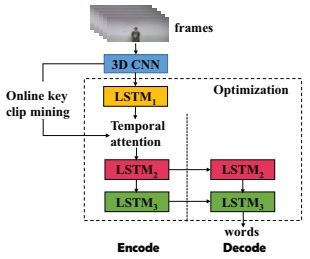} 
\end{center}
   \caption{An overview of the model proposed by Guo et al. \cite{Guo-2018}: A hybrid model including the combination of a 3D Convolutional Neural Network (3DCNN) and a LSTM-based \cite{LSTM,LSTM-org} encoder-decoder to translate from sign videos to text outputs.}
\label{Fig 5}
\end{figure}

\begin{figure}[t]
\begin{center}
\includegraphics[width=1\linewidth]{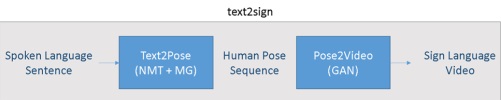} 
\end{center}
   \caption{An overview of the model proposed by Stoll et al. \cite{stoll2020}: A hybrid model to automatic SLP using NMT, GANs, and motion generation.}
\label{Fig 6}
\end{figure}

Regarding the sign language, Camgoz et al. proposed a combination of a seq2seq model with a CNN model to translate sign videos to spoken language sentences \cite{Camgoz-2018}. They used an attention-based Encoder-Decoder network with the attention weights defined as follows:
\[ \gamma_{n}^{u} = \frac{exp(score(h_{u},o_{n}))}{\sum_{n^{'} = 1}^{N}exp(score(h_{u},o_{n^{'}}))}  (7)
\]
where $h_{u},o_{n}, N$, are the hidden state, output, and sequence length, respectively. While results on the first continuous sign language translation dataset, PHOENIX14T, showed promising results, it would be interesting to extend the attention mechanisms to the spatial domain to align building blocks of signs with their spoken language translations. In another work, Guo et al. \cite{Guo-2018} designed a hybrid model including the combination of a 3D Convolutional Neural Network (3DCNN) and an LSTM-based \cite{LSTM,LSTM-org} encoder-decoder to translate from sign videos to text outputs (See Figure \ref{Fig 5}). Results on their own dataset showed a 0.071 \%\ improvement margin of the precision metric compared to state-of-the-art models. However, the unseen sentence translation is still a challenging problem with limited sentence data. Dilated convolutions and Transformer are two approaches that are also used for sign language translation \cite{Kalchbrenner,Vaswani}. Stoll et al. \cite{stoll2020} proposed a hybrid model to automatic SLP using NMT, GANs, and motion generation. The proposed model generates sign videos from spoken language sentences with a minimal level of data annotation for training. This model first translates spoken language sentences into sign pose sequences. Then, a generative model is used to generate the plausible sign language videos. Results on the PHOENIX14T Sign Language Translation dataset show comparable results compared to state-of-the-art alternatives (See Figure \ref{Fig 6}).

While NMT-based methods achieved successful results in translation tasks, some major challenges need to be solved. Domain adaptation is the first challenge in this area. Since the translation between different domains is affected by different rules, domain adaptation is a crucial requirement in developing machine translation systems targeted to a specific use case. The second challenge is regarding the amount of training data. Especially in deep learning-based models, increasing the amount of data can lead to better results. Another difficulty is dealing with uncommon words. The translation models perform poorly on these words. Words alignment and adjusting the beam search parameters are the other challenges for NMT-based models. The promising results of current deep learning-based models set an underpin for future research in this area.

\subsubsection{Motion Graph approaches}
\begin{figure}[t]
\begin{center}
\includegraphics[width=1\linewidth]{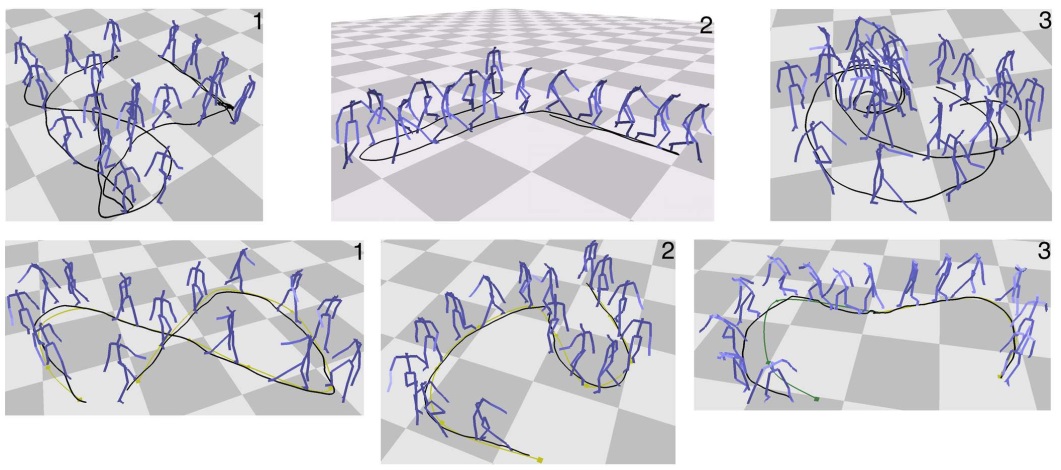} 
\end{center}
   \caption{An overview of the model proposed by Kovar et al. \cite{Kovar}: A general framework for extracting particular graph walks that satisfy the users specifications.}
\label{Fig 7}
\end{figure}

Motion Graph (MG), as a computer graphic method for dynamically animating characters, is defined as a directed graph constructed from motion capture data. MG can generate new sequences to satisfy specific goals. In SLP, MG can be combined with an NMT-based network to make a continuous-text-to-pose translation. One of the early efforts on MG is dated back to 2002, where a general framework was proposed by Kovar et al. \cite{Kovar} for extracting particular graph walks that satisfy the user’s specifications. Distance between two frames was defined as a distance between two point clouds (See Figure \ref{Fig 7}). To make the transitions, alignment and interpolation of the motions and positions were used between the joints. The total error, $f (w)$ of a path in the model is defined as follows: 
\[ f(w) = f([e_{1},...,e_{n}]) = \sum_{i=1}^{n}{g([e_{1},...e_{i-1}],e_{i})}    (8)   \],
where $g$ is a scalar function that evaluates the additional error accrued by appending an edge $e$ to the existing path $w$, which may be the empty path. Finally, the branch and bound search algorithm was applied to the graph.

In another work, Arikan and Forsyth \cite{Arikan} used the joint positions, velocities, and accelerations parameters to define the distance between two consecutive frames. Given
a sequence of edges $e_{1},...,e_{n}$, a score is assigned to each path using the following function:
\[ S(e_{1},...,e_{n}) = w_{c} * \sum_{i=1}^{n}{cost(e_{i})} +\] 
\[{w_{f}*F + w_{b}*B + w_{j}*J}  (9) \]
where $w_{c}, w_{f}, w_{b}, w_{j}$ are weights for the quality (continuity) of the motion, how well the length of the motion is satisfied, how well the body constraints are satisfied and how well the joints constraints are defined. $F$ is the squared difference between the actual and the required number of frames. $B$ is the squared distance between the actual and the required position and orientation of the constraint. $J$ is the squared distance between the actual and the required position of the constraint. In addition, the discontinuity between two clips was calculated using a smoothing function. After summarizing the graph, the random search was applied to the graph. 

\begin{figure}[t]
\begin{center}
\includegraphics[width=1\linewidth]{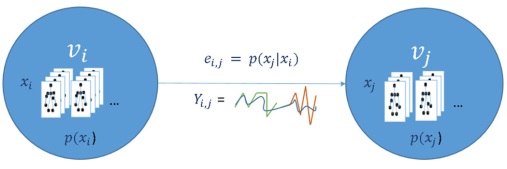}   
\end{center}
   \caption{An overview of the graph nodes in a model proposed by Stoll et al. \cite{stoll2020} for SLP. Each node contains one or more motion primitives and a prior distribution. The transition probability between two nodes is defined
as the probability of motion primitive.}
\label{Fig 8}
\end{figure}

A two-layer representation of motion data in another approach was proposed by Lee et al. \cite{Lee}. In the first layer, data was modeled as a first-order Markov process and the transition probabilities were calculated using the distances of weighted joint angles and velocities. A cluster analysis was performed on the second layer, namely cluster forest, to generalize the motions. The proposed hierarchical motion representation approach adapts the existing motion of a human-like character to have desired features included by a set of constraints. Results confirm the relative improvement by employing a curve fitting technique that minimizes a local approximation error.
A continuous SLP model has been proposed by Stoll et al. \cite{stoll2020} using pose data. The sign glosses were embedded to an MG with the transition probabilities provided by an NMT decoder at each time step (See Figure \ref{Fig 8}).

Although MG can generate plausible and controllable motion through a database of motion capture, it faces some challenges. The first challenge is regarding access to data. To show the model potential with a truly diverse set of actions, a large set of data is necessary. The scalability and computational complexity of the graph to select the best transitions are the other challenges in MG. Furthermore, since the number of edges leaving a node increases with the size of the graph, the branching factor in the search algorithm will increase as well. 

\subsubsection{Conditional image/video generation}
The field of automatic image/video generation has experienced a remarkable evolution in recent years. However, the task of video generation is challenging since the content between consecutive frames has to be consistent, showing a plausible motion. These challenges are more difficult in SLP due to the need for human video generation. The complexity and variety of actions and appearances in these videos are high and challenging. Controlling the content of the generated videos is crucial yet difficult.

With the recent advances in deep learning, the field of automatic image/video generation has seen different approaches employing neural network-based architectures, such as CNNs \cite{Chen,Oord}, RNNs \cite{Gregor,Oord-pixelRNN}, Variational Auto-Encoders (VAEs) \cite{VAE}, conditional VAEs \cite{CVAE}, and GAN \cite{GAN}. VAEs and GANs are generally combined to benefit from the VAE’s stability and the GAN’s discriminative nature. Most relevant to SLP, a hybrid model, including a VAE and GAN combination, has been proposed to image generation of people \cite{Ma-2017,Siarohin} and video generation of people performing sign language \cite{Stoll-sign,Vasani}. Furthermore, there are some models for image/video generation that can be used in SLP. For example, Chen and Koltun \cite{Chen-20188} proposed a CNN-based model to generate photographic images given semantic label maps. Van den Oord et al. \cite{Oord} proposed a deep learning-based model, namely PixelRNNs, to sequentially generate the image pixels along the two spatial dimensions. Gregor et al. \cite{Gregor} developed an RNN-based architecture, including an encoder and a decoder network to compress the real images presented during training and refine images after receiving codes. Karras et al. \cite{Karras} designed a deep generative model, entitled StyleGAN, to adjust the image style at each convolution layer. Kataoka et al. \cite{Kataoka} proposed a model using the combination of GAN and attention mechanism. Benefiting from the attention mechanism, this model can generate images containing high detailed content.

While deep learning-based generative models have recently achieved remarkable results, there exist major challenges in their training. Mode collapse, non-convergence and instability, suitable objective function, and optimization algorithm are some of these challenges. However, several strategies have been recently proposed to address a better design and optimization of them. Appropriate design of network architecture, proper objective functions, and optimization algorithms are some of the proposed techniques to improve the performance of deep learning-based models.

\subsubsection{Other models}
\begin{figure}[t]
\begin{center}
\includegraphics[width=1\linewidth]{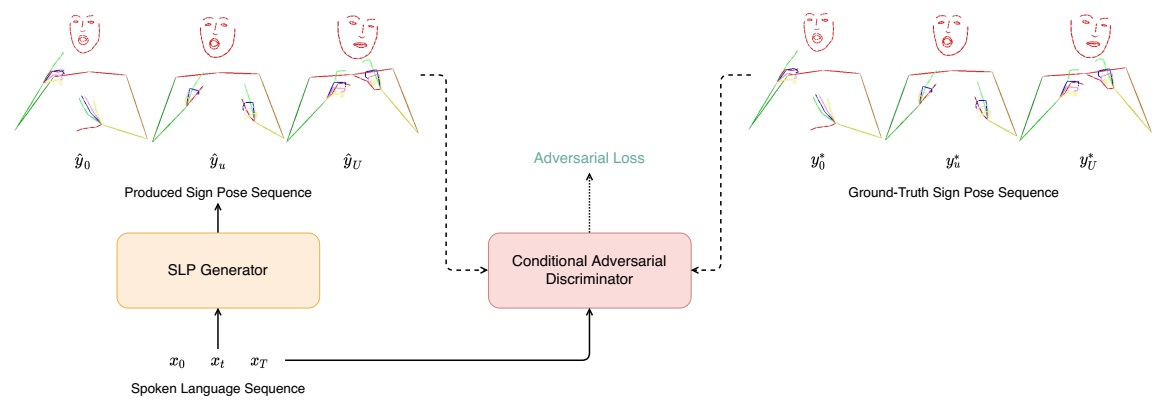}   
\end{center}
   \caption{An overview of the model proposed by Saunders et al.\cite{SaundersBMVC}. In this model, a Conditional Adversarial Discriminator measuring the realism of Sign Pose Sequences produced by an SLP Generator.}
\label{Fig 9}
\end{figure}

In addition to the previous categories, some models have been proposed to SLP using different deep learning models. For example, Saunders et al. \cite{SaundersBMVC} proposed a Progressive Transformers, as a deep learning-based model, to generate continuous sign sequences from spoken language sentences (See Figure \ref{Fig 9}). They formalized the model training process as an adversarial training scheme using a minimax game. To this end, the generator, $G$, aims to minimize the following equation, whilst D maximizes it:
\[\min_{G} \max_{D} \mathcal{L}_{GAN}(G,D) = \]
\[ [log D(Y^{*}|X)] + E[log(1-D(G(X)|X))]  (10) \]

where $Y^{*}$ and $G(X)$ are the ground truth and the produced sign pose sequences, respectively. Results on the PHOENIX14T dataset show the effectiveness of the proposed approach. However, the model needs to further increase the realism of sign production by generating photo-realistic human signers, Furthermore, user studies in collaboration with the Deaf are required to evaluate the reception of the produced sign pose sequences.

\begin{figure}[t]
\begin{center}
\includegraphics[width=11cm,height=6.5cm,keepaspectratio]{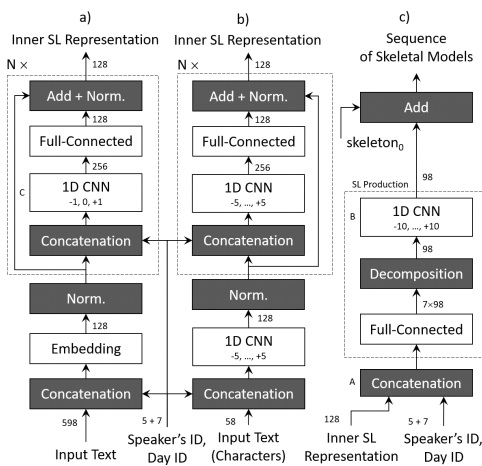}  
\end{center}
   \caption{Schematic diagram of a model proposed by Zelinka and Kanis \cite{Zelinka}, including three blocks: (a) the feed-forward model for a text-to-signs translation for word-level features, (b) the feed-forward model for a text-to-signs translation for character-level features, and (c) model for a sign-to-skeleton transformation.}
\label{Fig 10}
\end{figure}

In another work, Zelinka and Kanis \cite{Zelinka} designed a sign language synthesis system focusing on skeletal data production. A feed-forward transformer and a recurrent transformer, as deep learning-based models, along with the attention mechanism were used to enhance the model performance (See Figure \ref{Fig 10}). The loss of the proposed model for a sequence $a = (a_{1}, ..., a_{n_{a}})$ and a sequence $b = (b_{a}, ..., b_{n_{b}})$ is defined as follows:
\[ \varepsilon = \frac{\sum_{i=1}^{n_{a}} \sum_{j=1}^{n_{b}} w_{i,j} \|a_{i} - b_{j} \|_{D}^{2}}{\sum_{i=1}^{n_{a}} \sum_{j=1}^{n_{b}} w_{i,j}}     (11) \]

where $w(a,b) = [{w_{i,j}}]_{j = 1, ..., n_{b}}^{i = 1, ..., n_{a}}$ is an attention matrix and
${\|.\|}_{D}^{2}$ is a chosen metric.

Saunders et al. \cite{Saunders-arxiv} proposed a generative-based model to generate photo-realistic continuous sign videos from text inputs. They combined a transformer with a Mixture Density Network (MDN) to manage the translation from text to skeletal pose. The adversarial loss of the proposed model is defined as follows:
   \[ \mathcal{L}_{Total} = \min_{G}(( \max_{D_{i}} \sum_{i=1}^{k} \mathcal{L}_{GAN}(G,D_{i})) + \] \[ \lambda_{FM} \sum_{i=1}^{k} \mathcal{L}_{FM}(G,D_{i}) + \lambda_{VGG} \] \[ \mathcal{L}_{VGG}(G(y_{t},I^{S})) + \lambda_{KEY} \mathcal{L}_{KEY} (G,D_{H}) + \lambda_{T} \mathcal{L}_{T}(G))      (12) \]

Tornay et al. \cite{Tornay} designed an SLP assessment approach using multi-channel information (hand shape, hand movement, mouthing, facial expression). In this approach, two linguistic aspects are considered: the generated lexeme and the generated forms. Two loss functions corresponding to these linguistic aspects are calculated as follows:
\[
    \mathcal{S}_{lex} = \frac{1}{N} \sum_{n=1}^{N} \frac{\sum_{t=t_{n}^{b}}^{t_{n}^{e}}l(y_{n},z_{t})}{{t_{n}^{e}}-{t_{n}^{b}}+1}    (13)
\]

\[
    \mathcal{S}_{form}^{f} = \frac{1}{N} \sum_{n=1}^{N} \frac{\sum_{t=t_{n}^{b}}^{t_{n}^{e}} SKL(y_{n,f}z_{t,f})}{{t_{n}^{e}}-{t_{n}^{b}}+1}    (14)
\]

where $\mathcal{S}_{form}^{f}$ is the state duration normalized form-level score for each channel $f$, $\mathcal{S}_{lex}$ is the state duration normalized lexeme-level score, $l(y_{n},z_{t})$ is the local score defined by symmetric KL-divergence (SKL) between the probability distributions. $z$ and $y$ are the test sign production and the sequence of stacked categorical distributions corresponding to the KL-HMM representing the target reference lexeme. While results on the SMILE DSGS dataset show a promising lexeme and form levels assessment, they need to focus on assessment of a case that lexeme is correct but the form is incorrect.

Using the capabilities of different methods in this category has led to successful results in SLP. However, the challenge of the model complexity still remains an open issue. Making a trade-off between accuracy vs. task complexity is a key element.

\begin{table*}
\caption{\label{Table 3} A summary of the main characteristics of the reviewed models.}
\begin{center}
\begin{tabular}{p{3cm}p{6cm}p{3cm}}
\hline
\textbf{Query} & \textbf{Available choices} & \textbf{Most used}\\
\hline\hline
Methods & Avatar, NMT, Motion Graph, Image/video generation	& Image/video generation\\
Input modalities & Image (RGB), Skeleton, Video, Text, Speech & Text\\
Datasets & PHOENIX14T, Czech news, Own datasets & PHOENIX14T\\
Production modalities & Isolated, Continuous & Continuous\\
Architectures &	Static: GAN, AE, VAE & Static: GAN\\
& Dynamic: LSTM, GRU & Dynamic: LSTM\\
Generative models & AE, VAE, GAN & GAN\\
Evaluation metrics & Accuracy, Word Error Rate, BLEU, ROUGE	& BLEU\\
Features & Face, Hand, Body, Fused features	& Fused features\\
\hline
\end{tabular}
\end{center}
\end{table*}

\begin{table*}[h!]
\thispagestyle{empty}
\caption{\label{Table 4} Summary of deep SLP models.}
\begin{center}
{\small
 \noindent\begin{tabular}{p{0.8cm}p{0.7cm}p{1.5cm}p{2.2cm}p{1.8cm}p{8cm}}
 \hline \hline
 \textbf{Year} & \textbf{Ref} & \textbf{Feature} & \textbf{Input modality} &  \textbf{Dataset} & \textbf{Description}\\
 \hline \hline
 2011 & \cite{Kipp-Avatar} & Avatar & RGB video & ViSiCAST & \textbf{Pros.} Proposing a gloss-based tool focusing on the animation content evaluating using a new metric for comparing avatars with human signers.
\textbf{Cons.} Need to include non-manual features of human signers.\\
2016 & \cite{McDonald} & Avatar & RGB video & Own dataset & \textbf{Pros.}  Automatically adding realism to the generated images, low computational complexity.
\textbf{Cons.} Need to place the position of the shoulder and torso extension on the position of the avatar’s elbow, rather than the IK end-effector.\\
2016 & \cite{Gibet} & Avatar & RGB video & Own dataset & \textbf{Pros.} Easy to understand with high viewer acceptance of the sign avatars.
\textbf{Cons.} Limited to the small set of sign phrases.\\
2018 & \cite{Camgoz-2018} & NMT	& RGB video	& PHOENIX-Weather 2014T	& \textbf{Pros.} Robust to jointly align, recognize, and translate sign videos.
\textbf{Cons.} Need to align the signs in the spatial domain.\\
2018 & \cite{Guo-2018} & NMT & RGB video & Own dataset & \textbf{Pros.} Robust to align the word order corresponding to visual content in sentences.
\textbf{Cons.} Need to generalize to additional datasets.\\
2020 & \cite{stoll2020} & NMT, MG & Text	& PHOENIX14T & \textbf{Pros.} Robust to minimal gloss and skeletal level annotations for model training.
\textbf{Cons.} Model complexity is high.\\
2020 & \cite{Saunders}	& Others & Text	& PHOENIX14 & \textbf{Pros.} Robust to the dynamic length of output sign sequence.
\textbf{Cons.} Model performance can be improved including non-manual information.\\
2020 & \cite{Zelinka} & Others	& Text	& Czech news & \textbf{Pros.} Robust to the missing skeletons parts.
\textbf{Cons.} Model performance can be improved including information of facial expressions.\\
2020 & \cite{Saunders-arxiv} & Others & Text	& PHOENIX14T & \textbf{Pros.} Robust to non-manual feature production.
\textbf{Cons.} Need to increase the realism of the generated signs.\\
2020 & \cite{Camgoz} & Others & Text	& PHOENIX14T &	\textbf{Pros.} No need to the gloss information.
\textbf{Cons.} Model complexity is high.\\
2020 & \cite{Saunders-BMVC} & Others	& Text & PHOENIX14T	& \textbf{Pros.} Robust to manual feature production.
\textbf{Cons.} Need to increase the realism of the generated signs.\\
\hline
\noalign{\smallskip}
\end{tabular}
 }
\end{center}
\end{table*}

\section{General framework for SLP}
\begin{figure*}[t]
\begin{center}
\includegraphics[width=1\linewidth]{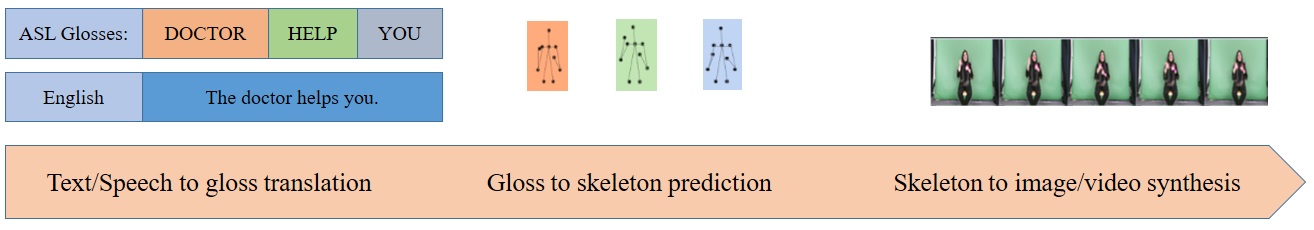}   
\end{center}
   \caption{ A general framework for SLP.}
\label{Fig 11}
\end{figure*}

SLP can be decomposed into some intermediate steps or addressed as an end-to-end translation task. As we reviewed in the previous sections, there are different translation models applicable in sign language. In this section, we present the common intermediate steps used in SLP (See Figure \ref{Fig 11}).

\textbf{Text/Speech to gloss translation: } Gloss is defined as written information of a sign word translated from the spoken language. It contains the facial and body grammar presented during the signing. For instance, let translate an English sentence, “I am Anna”, into sign language. To this end, we need to translate “I am” and “Anna” separately but finger-spelling for a letter-by-letter translation corresponding to “Anna” is needed. Finally, we have this: “EM FS-ANNA”, where “FS” denotes the start of a finger-spelling sequence. While the gloss is not a correct translation, it can provide suitable spoken language morphemes containing some conceptual information of the signs. The process of the spoken to gloss translation can be seen as a sequence-to-sequence task. In this task, various models from speech recognition and NMT, especially Deep Learning-based models, can be employed.

\textbf{Gloss to skeleton prediction: } This step aims to generate the human pose information corresponding to the sign gloss sequences. To this end, different parts of the human pose, including accurate finger locations, arm and torso position, and facial expressions, are considered. Like the previous step, this step can benefit from recent developments in Deep Learning. Attention-based models are one of the effective techniques employed for mapping from the textual input to the skeleton sequences.

\textbf{Skeleton to image/video synthesis: } Two general approaches are used in this step: animating an avatar and generating video frames. In the first approach, the skeleton keypoints are used to animate an avatar. Motion smoothing and interpolation are two techniques used before final rendering. While the video generation from the skeleton keypoints is hard, recent improvements in the Deep Learning-based skeleton-to-video translation are promising.

\section{Performance evaluation}
In this section, the results of the previously analyzed SLP models on the most popular datasets are presented. 

\subsection{Evaluation metrics and protocols}
Generally, the evaluation metrics measure the output quality by comparing the system output against the ground truth output corresponding to the source data. In SLP, the visual/lingual evaluation metrics are used to evaluate the correctness of the generated visual/lingual outputs: 

\textbf{Visual evaluation metrics:} To evaluate the quality of the generated sign image/video, the Structural Similarity Index Measurement (SSIM) \cite{Wang2018b}, Peak Signal-to-Noise Ratio (PSNR), and Mean Squared Error (MSE), as three well-known metrics for assessing image quality, are used in the proposed models for SLP. SSIM actually measures the perceptual difference between two images. In SLP, this metric is used to compare the generated synthetic image to its ground truth image. PSNR and MSE are metrics used to assess the quality of compressed images compared to their original. In SLP, the MSE is used to calculate the average squared error between a synthetic image and its ground truth image. In contrast, PSNR measures the peak error in dB, using the MSE metric.

\textbf{Lingual evaluation metrics:} Some of the most familiar machine translation metrics like BLEU@N \cite{Bleu}, METEOR \cite{Meteor}, ROUGE \cite{Rouge}, CIDEr \cite{Cider} are used to evaluate the translation performance of the proposed models in SLP. These metrics have acceptable relevancy with human judgment. In the BLEU@N metric, the matched N-grams between the machine-generated and the ground truth answer are utilized to compute the precision score. BLEU@N metric is calculated for N = 1 to 4, where shorter N-grams are used to fulfill the adequacy and longer N-gram matching accounts for fluency. ROUGE-L is another machine translation metric that scores a machine-generated sentence using a recall-based criterion. CIDEr is a metric for evaluating machine-generated sentences using human consensus. 

\subsection{Results}
In this section, we report the quantitative results of the most relevant methods reviewed in the previous sections. We limited the quantitative results to the most common metrics and datasets. The results are compacted in one table, given that there are only a few works in SLP. ROUGE and BELU are the most used metrics for reporting the results of the model evaluation. As Table \ref{Table 5} shows, most of the proposed models for SLP are evaluated on the PHOENIX14T dataset. This dataset contains 8257 sequences being performed by 9 signers, which are annotated with both the sign glosses and spoken language translations. However, due to the limited number of signers in the dataset, it is necessary to use one or more large-scale datasets to train the generation network. Using multiple datasets is motivated by the fact that there is no single dataset that provides text-to-sign translations, a broad range of signers of different appearances, and high definition signing content. Using datasets from different subject domains and languages demonstrates the robustness and flexibility of the proposed methods, as it allows us to transfer knowledge between specialized datasets. This makes the approach suitable for translating between different spoken and signed languages, as well as other problems, such as text-conditioned image and video generation.\\
Currently, the proposed SLP systems cannot compete with existing avatar approaches. A large amount of high-resolution training data is necessary to obtain results comparable with motion capture and avatar-based approaches. However, the avatar-based approaches need detailed annotations using task-specific transcription languages, which can only be provided by expert linguists. Animating the avatar itself often involves a remarkable amount of hand-engineering. Motion capture-based approaches require high-fidelity data, which needs to be captured, cleaned, and stored at remarkable cost, decreasing the amount of data available, therefore, making this approach un-scalable. Given that recent approaches use automatic feature extraction methods, We think that in time these approaches will enable highly realistic, and cost-effective translation of spoken languages to sign languages, improving equal access for the Deaf and Hard of Hearing. Generating the high resolution and signer independent videos with signers of arbitrary appearance makes room to provide the highly realistic, expressive, and end-to-end SLP systems applicable in real-world communications. Additionally, developing stronger data-processing strategies to pay attention to the intricate features of sign language data, such as the size of motion and speed, can be effective.

\begin{table*}[h!]
\thispagestyle{empty}
\caption{\label{Table 5} Results of SLP models.}
\begin{center}
{\small
 \noindent\begin{tabular}{p{2cm}p{0.6cm}p{0.8cm}p{1cm}p{1cm}p{1cm}p{1cm}p{1cm}p{1cm}p{1cm}p{0.7cm}p{0.7cm}p{0.7cm}}
 \hline \hline
 \textbf{Model} & \textbf{Acc} & \textbf{CIDEr} & \textbf{ROUGE} & \textbf{METEOR} & \textbf{ WER} & \textbf{BLEU-1} & \textbf{BLEU-2} & \textbf{BLEU-3} & \textbf{BLEU-4} & \textbf{MSE} & \textbf{SSIM} & \textbf{FID} \\
 \hline \hline

S2G2T \cite{S2G2T} &-&-& 43.80 &-&-& 43.29 & 30.39 & 22.82 & 18.13 &-&-&\\			
HLSTM-attn \cite{Guo-2018} & 0.506 & 0.605 &-& 0.205 & 0.641 & 0.508 &	0.330 & 0.207 &-&-&-&-\\ 				
Text2Gloss \cite{stoll2020} &-&-& 48.10 &-& 4.53 & 50.67 & 32.25 & 21.54 & 15.26 &-& 0.727 & 64.01\\
Symbolic Transformer \cite{Saunders} &-&-& 54.55 &-&-& 55.18 & 37.10 & 26.24 & 19.10 &-&-&- \\
Progressive Transformer \cite{Saunders} &-&-& 32.02 &-&-& 31.80	& 19.19 & 13.51 & 10.43 &-&-&- \\			
NSLS \cite{Zelinka} &-&-&-&-&-&-&-&-&-& 11.94 &-&- \\	
SIGNGAN \cite{Saunders} &-&-& 29.05 &-&-& 27.63 & 19.26 & 14.84	& 12.18 &-& 0.759 & 27.75 \\
EDN \cite{EDN} &-&-&-&-&-&-&-&-&-&-& 0.737	& 41.54 \\
vid2vid \cite{vid2vid} &-&-&-&-&-&-&-&-&-&-& 0.750 & 56.17 \\
Pix2PixHD \cite{Pix2PixHD} &-&-&-&-&-&-&-&-&-&-& 0.737 & 42.57 \\
\hline
\noalign{\smallskip}
\end{tabular}
 }
\end{center}
\end{table*}

\section{Discussion}
In this survey, we presented a detailed review of the recent advancements in SLP. We presented a taxonomy that summarized the main concepts related to SLP. We categorized recent works in SLP providing separate discussions in each category. The proposed taxonomy covered different input modalities, datasets, applications, and proposed models. Here, we summarize the main findings:

\textbf{Input modalities:} Generally, vision and language modalities are two input modalities in SLP. While the visual modality includes the captured image/video data used in the training, the linguistic modality contains the text input from natural language, which is applicable in both the training and testing of the proposed models. Both categories benefit from deep learning approaches to improve model performance. RGB and skeleton are two common types of visual input data used in SLP models. While RGB images/videos contain high-resolution content, skeleton inputs decrease the parameter complexity of the model and assist in making a low-complex and fast model. GAN and LSTM are the two most used deep learning-based models in SLP for visual inputs. While successful results were achieved using these models, more effort is necessary to generate more lifelike and high-resolution sign images/videos acceptable by the Deaf community. Among the deep learning-based models for lingual modality, the NMT model is the most used model for input text processing. Other Seq2Seq models, such as RNN-based models, proved their effectiveness in many tasks. While accurate results were achieved using these models, more effort is necessary to overcome the existing challenges in the translation task, such as domain adaptation, uncommon words, words alignment, and word tokenization.

\textbf{Datasets:} The lack of a large annotated dataset is one of the major challenges in SLP. The collection and annotation of sign language data is an expensive task that needs the collaboration of linguistic experts and native speakers. While there are some publicly available datasets for SLP \cite{Athitsos,Bungeroth,Necati2018,Caselli,Duarte,Ko,Matthes}, they suffer from weakly annotated data for sign language. Furthermore, most of the available datasets in SLP contain a restricted domain of vocabularies/sentences. To make a real-world communication between the Deaf and hearing communities, access to a large-scale continuous sign language dataset, segmented on the sentence level, is necessary. In such a dataset, a paired form of the continuous sign language sentence and the corresponding spoken language sentence needs to be included. Just a few datasets meet these criteria \cite{Camgoz-2018,Duarte,Ko,Zelinka}. The point is that most of the aforementioned datasets cannot be used for end-to-end translation  \cite{Camgoz-2018,Ko,Zelinka}. Two public datasets, RWTH-Phoenix-2014T and How2Sign, are the most used datasets in SLP. The former includes German sign language sentences that can be used for text-to-sign language translation. The latter is a recently proposed multi-modal dataset used for speech-to-sign language translation. Though RWTH-PHOENIX-Weather 2014T \cite{Camgoz-2018} and How2Sign \cite{Duarte-dataset} provided the appropriate SLP evaluation benchmarks, they are not enough for the generalization of the SLP models. Furthermore, these datasets only include German and American sentences. Translating from the spoken language to a large diversity of sign languages is a major challenge for the Deaf community.

\textbf{Applications:} American Sign Language (ASL) is the most-used sign language in developed applications for SLP. Since it may be hard for Deaf people to read or write the spoken language, they need some tools for communication with other people in society. Furthermore, many interesting and useful applications on the Internet are not accessible for the Deaf community. To tackle these challenges, some projects have been proposed aiming to develop such tools. While these applications successfully made a bridge between Deaf and hearing communities, we are still far from having applications involving large vocabularies/sentences from complex real-world scenarios. One of the main challenges for these applications is a license right for usage. Another challenge is regarding the application domain. Most of these applications have been developed for very specific domains such as clinics, hospitals, and police stations. Improving the amount of available data and its quality can benefit the creation of these needed applications. Furthermore, understanding the Deaf culture is helpful to create systems that align with user needs and desires.

\begin{figure}[h]
\begin{center}
\includegraphics[width=1\linewidth]{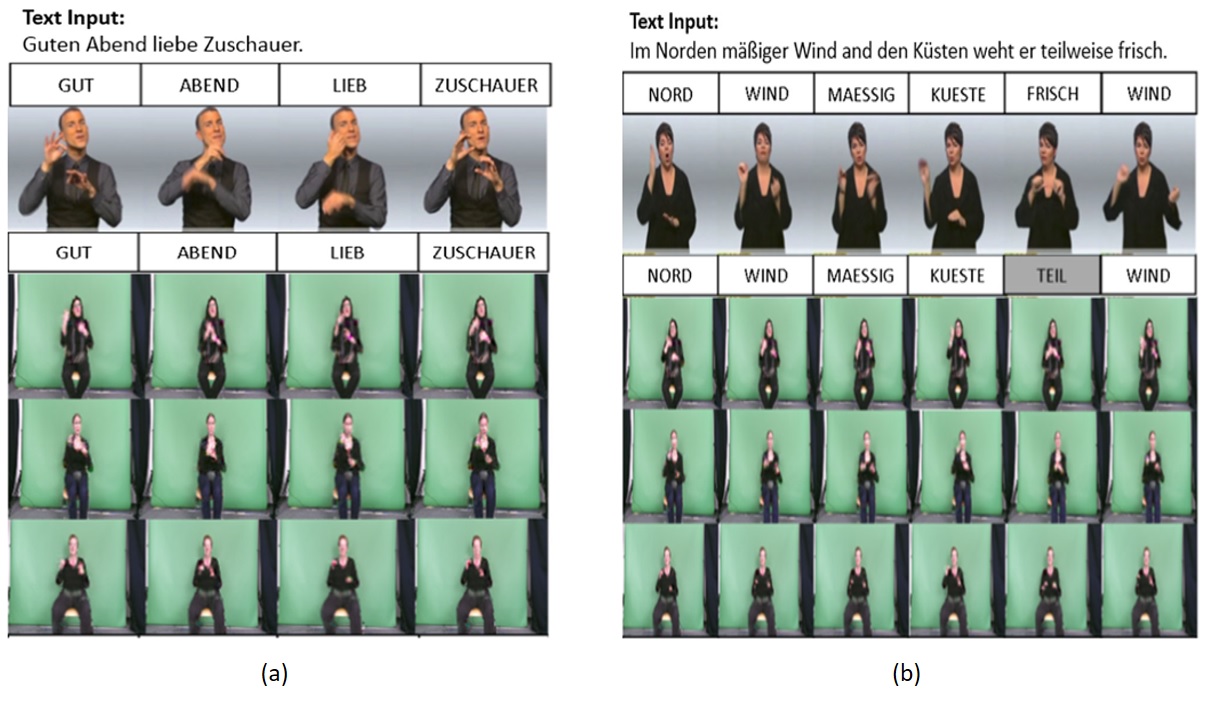} 
\end{center}
   \caption{Translation results from \cite{stoll2020}: (a) “Guten Abend liebe Zuschauer”. (Good evening dear viewers), (b) “Im Norden maessiger Wind an den Kuesten weht er teilweise frisch”. (Mild winds in the north, at the coast it blows fresh in parts).  Top row: Ground truth gloss and video, Bottom row: Generated gloss and video. This model combines an NMT network and GAN for SLP.}
\label{Fig 12}
\end{figure}

\begin{figure}[h]
\begin{center}
\includegraphics[width=1\linewidth]{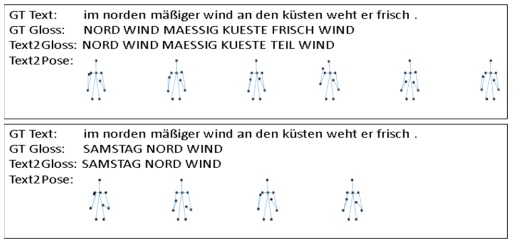}   
\end{center}
   \caption{Translation results from \cite{stoll2020}: Text from spoken language is translated to human pose sequences.}
\label{Fig 13}
\end{figure}

\begin{figure}[h]
\begin{center}
\includegraphics[width=1\linewidth]{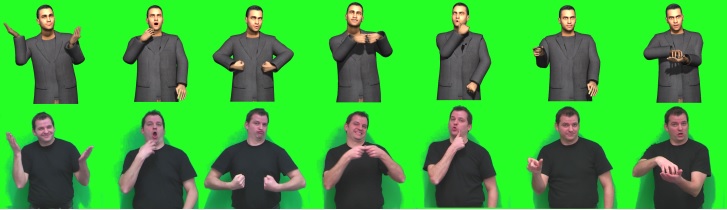}   
\end{center}
   \caption{Translation results from \cite{Kipp-Avatar}: A signing avatar is created using a character animation system. Top row: signing avatar, Bottom row: original video.}
\label{Fig 14}
\end{figure}

\textbf{Proposed models:} The proposed works in SLP can be categorized into five categories: Avatar approaches, NMT approaches, MG approaches, Conditional image/video generation approaches, and other approaches. Table \ref{Table 2 } shows a summary of state-of-the-art deep SLP models. Some samples of the generated videos and gloss annotations are shown in Figure \ref{Fig 12}, \ref{Fig 13}, and \ref{Fig 14}. Using the data collected from motion capture, avatars can be more usable and acceptable for reviewers. Avatars achieve highly realistic results but the results are restricted to a small set of phrases. This comes from the cost of the data collection and annotation. Furthermore, avatar data is not a scalable solution and needs expert knowledge to be inspected and polished. To cope with these problems and improve performance, deep learning-based models are used.

While NMT-based methods achieved significant results in translation tasks, some major challenges needs to be solved. Domain adaptation is the first challenge in this area. Since the translation in different domains need different styles and requirements, it is a crucial requirement in developing machine translation systems targeted at a specific use case. The second challenge is regarding the amount of available training data. Especially in deep learning-based models, increasing the amount of data can lead to better results. Another challenge is regarding to the uncommon words. The translation models perform poorly on these words. Words alignment and adjusting the beam search parameters are other challenges in NMT-based models.\\
Although MG can generate plausible and controllable motion through a database of motion capture, it faces some challenges. The first challenge is regarding limited access to data. To show the model potential with a truly diverse set of actions, a large set of data is necessary. The scalability and computational complexity of the graph to select the best transitions are other challenges in MG. Furthermore, since the number of edges leaving a node increases with the graph size, the branching factor in the search algorithm will increase as well. To automatically adjust the graph configuration and rely on the training data, instead of user interference, Graph Convolutional Network (GCN) could be used along with some refining algorithms to adopt the graph structure monotonically.

While GANs have recently achieved remarkable results for image/video generation, there exist major challenges in the training of GANs. Mode collapse, non-convergence and instability, suitable objective function, and optimization algorithm are some of these challenges. However, several suggestions have been recently proposed to address the better design and optimization of GANs. Appropriate design of network architecture, proper objective functions, and optimization algorithms are some of the proposed techniques to improve the performance of GAN-based models. Finally, the challenge of the model complexity still remains for hybrid models.

\textbf{Limitations:} In this survey, we presented recent advances in SLP and related areas using deep learning. While successful results have been achieved in SLP by recent deep learning-based models, there are some limitations that need to be addressed. The main challenge is regarding the Multi-Signer (MS) generation that is necessary for providing real-world communication in the Deaf community. To this end, we need to produce multiple signers of different appearances and configurations. Another limitation is the possibility of high-resolution and photo-realistic continuous sign language videos. Most of the proposed models in SLP can only generate low-resolution sign samples. Conditioning on human keypoints extracted from training data can decrease the parameter complexity of the model and assist to produce a high-resolution video sign. However, avatar-based models can successfully generate high-resolution video samples, though they are complex and expensive. In addition, the pruning algorithms of MG need to be improved by including additional features of sign language, such as duration and speed of motion.

\textbf{Future directions:} While recent models in SLP presented promising results relying on deep learning capabilities, there is still much room for improvement. Considering the discriminative power of self-attention, learning to fuse multiple input modalities to benefit from multi-channel information, learning structured spatio-temporal patterns (such as Graph Neural Networks models), and employing domain-specific prior knowledge on sign language are some possible future directions in this area. Furthermore, there are some exciting assistive technologies for Deaf and hearing-impaired people. A brief introduction to these technologies can get an insight to the researchers in SLP and also make a bridge between them and the corresponding technology requirements. These technologies fall into three device categories: hearing technology, alerting devices, and communication support technology. For example, let imagine a technology that assists a Deaf person go through a musical experience translated into another sensory modality. While the recent advances in SLP are promising, more endeavor is indispensable to provide a fast processing model in an uncontrolled environment considering rapid hand motions. It is clear that technology standardization and full interoperability among devices and platforms are prerequisites to having real-life communication between the hearing and hearing-impaired communities. Finally, since providing the data annotation is also challenging, recently some efforts have been done by Rastgoo et al. \cite{ZSL1}\cite{ZSL2} to overcome the annotation bottleneck. To this end, Zero-Shot Learning (ZSL) is employed for SLR. Using this approach, we hope to get closer to the real and accurate systems for bidirectional communication between Deaf and hearing people in society.\\

\section*{Acknowledgements}
This work has been partially supported by the HIS Company and Institute for Research in Fundamental Sciences (IPM) in Iran, Spanish project PID2019-105093GB-I00 (MINECO/FEDER, UE), and CERCA Programme/Generalitat de Catalunya, and ICREA under the ICREA Academia programme.


\end{document}